\documentclass[lettersize,journal]{IEEEtran}
\usepackage{amsmath,amsfonts}
\usepackage{algorithmic}
\usepackage{algorithm}
\usepackage{array}
\usepackage[caption=false,font=normalsize,labelfont=sf,textfont=sf]{subfig}
\usepackage{textcomp}
\usepackage{stfloats}
\usepackage{url}
\usepackage{verbatim}
\usepackage{graphicx}
\usepackage{cite}
\usepackage[justification=centering]{caption}
\usepackage{xcolor}
\usepackage{booktabs}
\usepackage{makecell}

\DeclareMathOperator{\expectation}{\mathop{\mathbb{E}}}

\usepackage{mathtools}
\DeclarePairedDelimiterX{\infdivx}[2]{(}{)}{%
  #1\;\delimsize\|\;#2%
}

\begin{document}

\title{Reinforcement Learning for Generative AI: A Survey}


\author{Yuanjiang Cao, Quan Z. Sheng, \IEEEmembership{Member,~IEEE}, 
Julian McAuley, Lina Yao, \IEEEmembership{Senior Member,~IEEE}, 

\thanks{
Yuanjiang Cao and 
Quan Z. Sheng (Michael Sheng) are with Department of Computing, Macquarie University, Sydney, NSW, AUS. Email: yuanjiang.cao@mq.edu.au, michael.sheng@mq.edu.au
}
\thanks{Julian McAuley is with University of California San Diego, California, USA. Email:jmcauley@eng.ucsd.edu}
\thanks{Lina Yao is with CSIRO's Data61 and University of New South Wales. Email: lina.yao@data61.csiro.au}
} 

\markboth{Journal of \LaTeX\ Class Files,~Vol.~14, No.~8, August~2021}%
{Shell \MakeLowercase{\textit{et al.}}: A Sample Article Using IEEEtran.cls for IEEE Journals}

\maketitle

\begin{abstract}
Deep Generative AI has been an essential topic in the machine learning community for a long time, and it can impact a number of application areas like text generation and computer vision. The major paradigm for training a generative model is maximum likelihood estimation. This formulation successfully establishes the objective of generative tasks, while it cannot satisfy all the requirements that a user might expect from a generative model. Reinforcement learning has demonstrated its power and flexibility to inject new training signals such as human inductive bias to build a performant model. Thereby, reinforcement learning has become a trending research field and has stretched the limits of generative AI in both model design and application. It is reasonable to summarize advances in recent years with a comprehensive review. Although there have been surveys in different application areas recently, this survey aims to shed light on a high-level review that spans a range of application areas. We provide a rigorous taxonomy and make sufficient coverage on various models and applications, including the fast-developing large language model area. We conclude this survey by showing the potential directions that might tackle the limit of current models and expand the frontiers for generative AI.
\end{abstract}


\section{Introduction}
Recent years have witnessed tremendous progress in generative AI, like variational autoencoders\cite{kingma2013auto}, autoregressive models\cite{van2016pixel}, adversarial generative nets\cite{goodfellow2014generative}, diffusion models\cite{ho2020denoising}, energy-based models\cite{collobert2008unified} and normalizing flows\cite{rezende2015variational}. The advancement of these models has brought the development of a broad range of applications, from neural language processing to image generation and scientific research.
Particularly, the emergence of Large Language Models (LLM), like ChatGPT\cite{openai2022chatgpt}, has changed the paradigm of industry and academia regarding how to develop the next generation of machine learning systems to bridge the gap toward general AI further.
Another fast-developing area is Diffusion models \cite{ho2020denoising}, which is the foundation for large models that generate high quality images, videos\cite{harvey2022flexible}, and medical image analysis\cite{song2021solving}. whose training requires large amounts of computing resources for performant image generation.
Generative models also power scientific research in molecular design and optimization. AlphaFold\cite{jumper2021highly} shows that protein structure can be effectively modeled by machine learning systems \cite{baek2021accurate,lin2023evolutionary}.

Training generative model is one of the cornerstones of generative AI research, which studies to design objective functions to guide the learning process. The major objective of generative models is Maximum Likelihood Estimation (MLE), in other words, decreasing the Kullback-Leibler(KL) divergence between generated distribution and target data distribution. However, in some cases, human want more than what the MLE can provide. Taking conditional text generation as an example, for a text generator, we hope not only that it achieves good performance on texts that exist in the training dataset, but also that it can output texts that satisfy other desired properties like diversity, coherence, human-like, and moral considerations. The discrepancy between evaluation metrics and training objectives can decrease the quality of generated outputs. These desired properties for a text generator expose that the gap between distribution fitting and desired properties makes the maximum likelihood objective insufficient. The generalization of models connects to the limit of the Negative Log Likelihood (NLL) objective as well. In some applications of generative AI, we hope the model can cope with out-of-distribution inputs or explores out-of-distribution. For example,  in novel molecule design, the goal of the learning process is to explore and generate unseen molecules instead of those in the dataset. A code summarizer or generator is expected to produce well-designed code for novel tasks instead of those in the dataset.

To address the aforementioned limits, the reinforcement learning 
has been proposed as a useful optional training paradigm to improve the performance of generation models.
Reinforcement learning is a training paradigm designed for learning from interaction \cite{sutton2018reinforcement}. It has flexible objectives in terms of the reward function, in contrast to the distribution modeling objective of supervised learning and unsupervised learning. Furthermore, many generation problems can be redefined as decision-making problems, creating the utility of RL methods on generation problems.

\begin{center}
\begin{figure*}[t]
\centering
\includegraphics[width=1.0\linewidth]{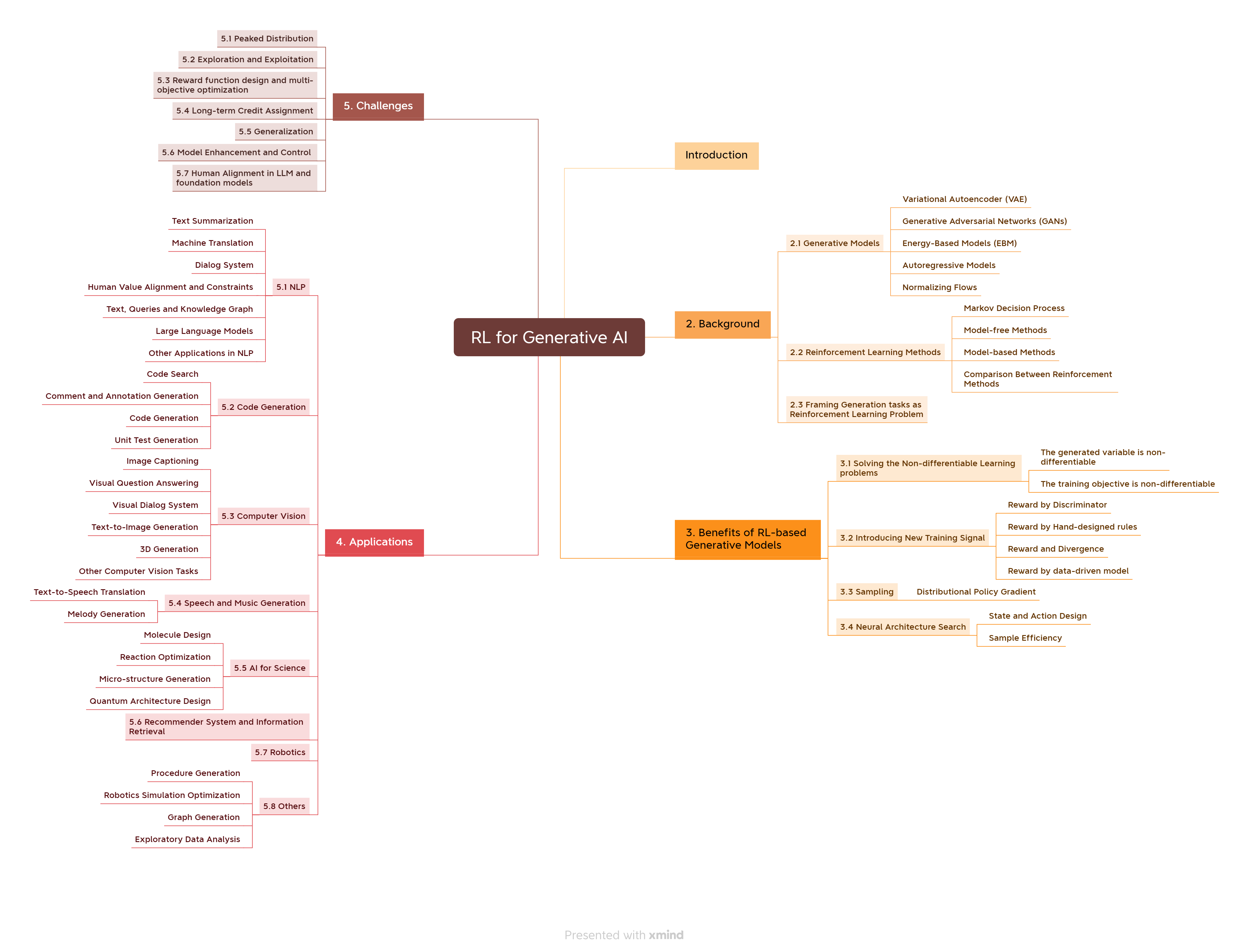}
\caption{The Overview Structure of This Survey}
\label{fig:rl-for-generative-ai}
\end{figure*}
\end{center}

\textbf{Why this survey}
There are numerous existing surveys and reviews of the application of reinforcement learning scattered in various models \cite{mohebbi2023games} and application areas, including neural language processing \cite{chen2017survey,de2021survey,uc2023survey,ni2023recent}, large language model \cite{sun2023reinforcement},
code generation \cite{zhang2022survey}, speech processing \cite{tan2021survey,lin2022reinforcement}, computer vision \cite{hossain2019comprehensive,le2022deep}, neural architecture search \cite{jaafra2019reinforcement,santra2021gradient}, and drug discovery \cite{du2022molgensurvey,fromer2023computer,bilodeau2022generative,tang2021generative,luukkonen2023artificial}. We have found one paper \cite{franceschelli2024reinforcement} which survey the area of RL applying on generative models, while it only investigates three perspectives in terms of objective design. In contrast, this work covers topics like using RL as a sampling method. For application, we explicitly list the application area, making readers easier to access topics of their interests and also inspire new ideas in the model discussion part. We also cover niche application areas like procedure generation and graph generation.

Our survey surpasses these previous surveys in terms of the selection criteria and the comprehensiveness of coverage in this area.
We not only summarize lines of research across multiple areas but also organize theoretical works that aim to improve generative models by reinforcement learning techniques. For instance, we cover the recent advancement that the efficiency of Diffusion Model could be enhanced by shortcut fine-tuning motivated by one of the reinforcement learning method: policy gradient, as shown in Section \ref{sec:benefits-new-signal}.

\textbf{Scope and Paper selection criteria.} This study offers a comprehensive analysis of the potential and obstacles associated with reinforcement learning, with a particular focus on deep reinforcement learning in the realm of generative AI. The analysis encompasses their intrinsic capabilities, 
 such as addressing non-differentiable learning issues, infusing generative AI with innovative training signals, and advances in sampling and neural architecture search. It also sheds light on the prevailing challenges, including peaked distribution, the conundrum of exploration versus exploitation, sparse reward scenarios, challenges in long-term credit allocation, and issues of generalization. Furthermore, the research delves into their practical applications across various domains, including Natural Language Processing (NLP), Computer Vision (CV), code synthesis, speech decoding, information extraction, recommendations, robotics, and AI's role in scientific endeavors. Emerging trends, such as the intricacies of reward function formulation, multi-objective optimization strategies, enhancing and controlling models, modeling human preferences, ensuring interoperability, and the integration of novel reinforcement learning techniques with Large Language Models (LLMs) and foundational models, are also meticulously explored.

In application areas, this survey mainly focuses on sequential generation, e.g. conditional text generation and code generation, but also contains less-mentioned vision tasks such as 3D point cloud completion. We select papers according to three selection criteria, impact, venues, and time. For classic models, we focus on high-quality works that are present in well-appreciated conferences and journals. For recent papers that push the boundaries like large language models, we relaxed the criteria. For some small branches, we list all the papers we can find.

We propose this survey to systematically review the application of reinforcement learning in generative AI, including models and applications. Our contributions include

\begin{itemize}
\item We perform a systematic and thorough examination of various directions within the field of generative models and problems using Reinforcement Learning. Recognizing the multifaceted nature of these areas, we carefully organize and present an exhaustive review that encompasses different methods, structures, and applications.

\item We developed a unified taxonomy, meticulously crafted to organize the extensive and varied literature in generative studies. This taxonomy not only classifies existing works according to common themes and methodologies but also draws attention to potential intersections and divergences within the field. 

\item With a specialized focus on Reinforcement Learning (RL) methodologies within generative AI, our research provides a detailed exploration of the contemporary challenges and avenues for development. We dissect the inherent complexities and hurdles in implementing RL methods and organize them in a manner that encapsulates the current state of the art. Furthermore, we identify several emerging directions that hold promise for innovative solutions.

\end{itemize}



\begin{center}
\begin{table*} 
\centering
\caption{Notations in this Survey}
\begin{tabular}{cc}
    \toprule
    Notation & Explanation \\
    \hline
    $x$, & the target random variable that a generative model aims to generate, \\
    $x_1, .., x_i, .., x_n$ & a sequence of variables, $x_i$ represents an element of $x$ when $x$ represents a sequence. This is useful in sequence modelling. \\
    $z$ & latent variable \\

    & \\

    $t$ & the time step in a sequence, often serve as an index\\
    $s_t$ & the state at time step $t$ in an MDP \\
    $a_t$ & the action at time step $t$ in an MDP \\
    $r_t$ & the reward given by the environment at time step $t$ in an MDP \\
    $\tau$ & a trajectory, a.k.a. a sequence of states, actions, and rewards $(s_0, a_0, s_1, r_1, ... s_n, r_n)$\\
    $R_t$ & $R_t=\sum_{i=t}^{T} \gamma^{i-t} r_i$ the discounted accumulative return over a trajectory \\
    $\gamma$ &, as shown abov, is the discount factor that decreases the impact of future rewards at an exponential rate \\
    $\pi$ & the policy of the RL agent \\
    $V_{\pi}(s_t)$ & the value function of an agent given a state at time step $t$ \\
    $Q_{\pi}(s_t, a_t)$ & the value function of an agent that takes a state and an action as input \\
    
    & \\
    $p(\cdot), q(\cdot)$ & the probability distribution of a given variable\\
    $\expectation_{x \sim p_{\mathit{x}}}[\cdot]$ & the expectation of some variable on the distribution of $x$ \\
    $D_{KL}(q\Vert p) $ & the KL Divergence between two distribution $p$ and $q$ on the same variable. \\

    & \\
    $D(\cdot)$ & the discriminator in GAN \\
    $G(\cdot)$ & the generator in GAN \\
    \bottomrule
\end{tabular}
\label{tab:taxonomy}
\end{table*}
\end{center}

\section{Preliminary and Background}
In this section, we will provide preliminaries about the relevant concepts and models, in terms of generative models and reinforcement learning. 

\subsection{Generative Models}
The topic of this survey is reinforcement learning applied in generative AI. Thus it is essential to provide a brief introduction to basic generative models, which lays the foundation for successive chapters. Before we dive into details, we list the most common notations throughout this survey in Table \ref{tab:taxonomy}.

Generation is a broad research area that scatters in various application areas. We take three examples for readers to understand what is generation, then we introduce the formal definition of key concepts.
The first example is a dialogue system like ChatGPT \cite{openai2022chatgpt}, one of the most popular generative AI that covers more than 180 million users \cite{anna2023exclusive}. If we want to build a dialogue system, we collect a dataset consists of pairs of user prompts and human response. The goal is to train a model that takes a sequence of prompts that consists of symbols and then return a sequence of symbols as response. The symbol comprises of alphabet characters, math symbols, other specific symbols. The second example is text-to-image generation application like Midjourney\cite{midjourney2024}, where users input a sequence of symbols and expect to get an image that satisfying the input. The third example is image-to-image translation. The user wants to transform a realistic image to other styles like an impressionistic image. The user needs to pipe the image and prompt, which is again a sequence of symbols, into the image generator. From above problem description, we can see that the goal of generation is typically requires to pipe the input variable $y$ into the model, which could be in the form of a sequential form $(y_1, y_2, ..., y_n)$ and the model emits a variable $x$, which could be represented in the form of a sequence  $(x_1, x_2, ..., x_m)$.

Given the input variable and output variable, it is easy to think that we can directly a probabilistic model $p(x|y)$ via machine learning methods. Interestingly, this is generally infeasible for the mismatch between data we have and the task requirement. For example, we want to build a dialogue system, however, most text data we can get is acquiring from webs, these texts are not dialogue data, which means it cannot be used to learn $p(x|y)$. 
Instead, we can train a model that models $p(x)$, the distribution of the data we can get, then we can adjust the model to the task we need. In the dialogue example, we can train a model with web dataset, then we can fine-tune the model based on the dialogue data we collected from interaction with human. For text-to-image generation, changing the learning objective from $p(x)$ to $p(x|y)$ is also not difficult. For simplicity, we describes generative models whose goal is to learn $p(x)$ in the following section.


Given a dataset $X$ composed of a set of samples $\{x^i| 1\leq i \geq n\}$, we aim to use a generative model $p(\hat{x})$ to generate a sample $\hat{x}$ that follow the true data distribution $p(x)$, where $x$ is the random variable that represents the true data sample, $n$ the number of samples in the dataset. The $x$ could be a scalar variable, a vector variable or a sequence of variables $(x_1, x_2, ..., x_n)$, where the subscript means the ordered index in the sequence, $x_i$ means one data point of the sequence.

\textbf{Training and Inference} The learning of a model contains two useful stages: training phase and inference stage. In training stage, a model is trained. Training a model requires an objective function or a loss function which reveals the goal the model optimizes towards. For example, language models typically use next word prediction as an objective. In this section, we introduces basic formulation of five generative models, and we use minimizing NLL as the objective function without additional description, which is supported by the maximum likelihood principle\cite{goodfellow2016deep}. The computation of NLL entails computing $p(x)$, therefore for all generative models in this subsection, the formulation is about how to compute $p(x)$.
In inference stage, the model should generate a sample $\hat{x}$ given a latent variable $z$ that is sampled from a given distribution like Gaussian Distribution. The inference stage is how a model is used to generate samples.

\textbf{Variational Autoencoder (VAE)}\cite{kingma2013auto,rezende2014stochastic} learns useful representations by reconstructing the input $x$ with a latent variable $z$ in consideration. Formally, we can decompose the distribution $p(x)$ by the latent variable:
\begin{align}
    \label{eq:generative-model-vae-integral}
    p(x) &= \int p(x|z)p(z) dz.  \\
    \label{eq:generative-model-vae} \ln p(x) &\geq D_{KL}(q(z|x) | p(z)) + \expectation_{q(z|x)} \ln p(x|z)  
\end{align}
But the integral in Equation \ref{eq:generative-model-vae-integral} is intractable in real-world applications. Therefore, an optional way is to approximate this conditional distribution by a simpler distribution $q(x)$ with an evidence lower bound (ELBO) in Equation \ref{eq:generative-model-vae}.
In inference stage, VAE could sample following Equation \ref{eq:generative-model-vae-integral}, sample a $z$ from Gaussian Distribution, then pipe $z$ into the decoder $p(x|z)$ to compute the distribution, then sample a $\hat{x}$ from the distribution.

\textbf{Generative Adversarial Networks (GANs)}\cite{goodfellow2014generative} are comprised of a discriminator and a generator. The discriminator is trained to classify where samples come from, real datasets or generated. The generator aims to trick the discriminator. Therefore, the two networks form a zero-sum game which consequently pushes the output distribution of the generator to approximate the real distribution. Formally, the objective of GAN can be defined as:
\begin{equation}
\label{eq:generative-model-gan}
    \min_{G} \max_{D} \expectation_{x \sim p_{\mathit{data}}(x)} D(x) + \expectation_{z \sim p_\mathit{g}(z)} [1 - D(G(z)) ]
\end{equation}
We can observe that generator could directly generates data samples by sampling a latent vector $z$ and then pipe the vector into the generator.

\textbf{Energy-Based Models (EBM)}\cite{lecun2006tutorial} represents any distribution density with an energy function by
\begin{equation}
\label{eq:generative-model-ebm}
    p(x) = \frac{e^{-E(x)}}{\int_{x^{'} \sim \mathcal{X}} e^{-E(x^{'})} }
\end{equation}
where $E(x)$ is an energy function that outputs a density of the a certain sample $x$. It could be better to understand the energy function as a compatibility\cite{lecun2006tutorial} when the energy function is used to model a joint distribution $E(x,y)$, the energy is lower when $x$ and $y$ are more compatible. For distribution $p(x)$, we could treat $E(x)$ as a density. EBMs do not pose constraint on tractability of the normalizing constant, making them more flexible\cite{song2021train}.
It can be seen from this formulation that the denominator is an integral over the space $\mathcal{X}$. If we want to evaluate the probability, we need to evaluate the energy function all $x$. If the space is too huge,  computing the probability is intractable.
This makes NLL objective infeasible. Instead, proxy objectives have been developed to optimize the EBM, a popular one is
\begin{align}
\label{eq:generative-model-ebm-proxy}
    \nabla_{\theta} \mathcal{L} &= \nabla_{\theta} -\operatorname{ln} p_{\theta}(x) \\
    &= \nabla_{\theta} E_{\theta}(x) + \nabla_{\theta} \operatorname{ln} \int_{x^{'} \sim \mathcal{X}} e^{-E_{\theta}(x^{'})} \\
    &\approx \nabla_{\theta} E_{\theta}(x) - \expectation_{x^{-} \sim p_{\theta}} \nabla_{\theta} E_{\theta}(x^-) 
\end{align}
where the integral term is approximated by a Markov Chain Monte Carlo (MCMC) method that samples from the EBM, forming a contrastive objective \cite{song2021train}.
Sampling from EBM is not trivial as well. Researchers tend to use MCMC for sampling \cite{song2021train}, fortunately, one line of research shows that it might take a few steps of sampling chain to get an acceptable sample\cite{carreira2005contrastive}.

\textbf{Autoregressive Models (AR)} decompose the probability distribution of a variable $x$ into a sequence of variables by the chain rule:
\begin{equation}
\label{eq:generative-model-autoregressive}
    p(x) = p(x_1, x_2, ..., x_n) = \prod_i^n p(x_i|x_1, ..., x_{i-1})
\end{equation}
It can be naturally applied to sequence generation tasks like text generation and molecule design. It is not so obvious that AR models could model an image by breaking the image into a sequence of patches, allowing AR to be used for image generation. The ordering of generation is fixed and cannot be changed during training and inference. The computational complexity of the sampling process is linear in the number of steps in the generation, and the sequential ordering makes parallel computation difficult which leads to less efficiency. 
The training objective could minimize the negative log likelihood of $p(x)$, and sampling process could generate tokens one by one, where tokens refer to $x_{i}$ in Equation \ref{eq:generative-model-autoregressive}.
Although it seems the sequential generation process prevent parallel computing of generation and limits the scalability of AR, \cite{vaswani2017attention,pope2023efficiently,freitag2017beam} etc. have been proposed to address the scalability and AR has become one of the most useful generative models.

\textbf{Normalizing Flows}\cite{rezende2015variational,dinh2014nice,kobyzev2020normalizing}
Besides autoregressive models, another representative generative model that is capable of direct optimization of NLL objective is Normalizing Flow. Given a latent variable $z$ of a tractable distribution $p(z)$, normalizing flow can sample a data point $x$ by transforming $z$ through an invertible function $f$, forming $x = f(z)$. Because the $f(\cdot)$ is an invertible function, the distribution of $x$ could be computed by 
\begin{equation}
\label{eq:normalizing-flows-z-x}
    p(x)=p(z) \left | \det \frac{\partial f(z)}{\partial z} \right |^{-1}
\end{equation}
In Equation \ref{eq:normalizing-flows-z-x}, normalizing flows use the change of variable rule of distributions to get the distribution of $x$. This formulation allows fast sampling and inference by $f(z)$. The constraint of invertible functions is strong that one step transformation is generally insufficient when the distribution of $x$ is complex, which leads to a combination of multiple invertible transformations:
\begin{equation}
\label{eq:generative-model-normalizing-flows}
    \ln p(x_n) = \ln p(x_1) - \sum_{i} ln \left | \det \frac{f_i}{x_{i-1}} \right |
\end{equation}
This formulation requires each transformation $x_{i} = f_{i}(x_{i-1})$ to efficiently compute Jacobian determinant as well as to be expressive and invertible.
Capability of tractable sampling and inference makes training and inference tractable. During training, normalizing flows can compute $p(x)$ directly by Equation \ref{eq:normalizing-flows-z-x}. The sampling process has been explained above.

\textbf{Diffusion Model} is a type of generative model that injects noise into the data and learn a reverse process to generate samples. Here we briefly introduce one widely adopted diffusion model, Denoising Diffusion Probabilistic Models (DDPM). A DDPM model consists of two Markov chains, a forward chain $q(x_{K},\dotsc,x_1|x_0)=\prod_{k=1}^{K} q(x_k|x_{k-1})$ that perturbs the data sample with noises, and a backward chain $p_{\theta}(x_0,x_1,\dotsc,x_K)==\prod_{k=K}^{1} q(x_{k-1}|x_{k})$ that recover a sample from the data distribution given a random noise sample. Note that for simplicity, we slightly abuse the subscription of $x$ as a series of perturbing or denoising steps $k\in [1,K]$. The $\theta$ means the reverse process is parameterized by a neural network. The neural network is trained by minimizing the KL divergence between forward process and reverse process.
\begin{equation}
\label{eq:generative-model-diffusion-model}
    D_{KL}(q\Vert p_{\theta}) \geq \expectation[- \operatorname{log} p_{\theta}(x_0)]
\end{equation}
where the KL divergence is a upper bound of the NLL objective.  Hu et al. links the KL divergence to the following loss function:
\begin{equation}
\label{eq:generative-model-diffusion-model-hu}
    \expectation_{k\sim [1,K], x_0\sim q(x_0), \epsilon \sim \mathcal{N}(0,I)} [\lambda(k) \left\Vert \epsilon - \epsilon_{\theta}(x_k, k) \right\Vert^2]
\end{equation}
where $\epsilon$ is the Gaussian noise used in forward computation and $\lambda(k)$ is a positive weighting function that is necessary to link KL divergence to this formulation.

\textbf{Large Language Model} is a language model that uses very large neural networks as models that typically have billons of parameters in order to train a generalized model for various applications. The famous ChatGPT\cite{openai2022chatgpt} is a chatbot powered by the large language models. They are typically trained by two steps, the first step using a next word prediction objective, and the second step to use various methods to finetune the model in order to constrain the behaviors of the large models.

\subsection{Reinforcement Learning Methods}
\label{sec:background-rl}

Reinforcement learning is a computational approach to automating goal-directed learning and decision-making \cite{sutton2018reinforcement}. 
In this section, we introduce the Markov Decision Process (MDP), a formulation that can be widely applied to reinforcement learning problems. After formally establishing the problem, we introduce the categorization of reinforcement learning methods and the key details of these methods.

\subsubsection{Markov Decision Process}
Markov Decision Process is a classical formalization of sequential decision making \cite{sutton2018reinforcement} where actions have impacts on subsequent states and rewards. 

The complete formulation model of an MDP contains the following five elements:
\begin{itemize}
\item $\mathcal{S}$ is a set of states of the environment
\item $\mathcal{A}$ is a set of actions of the agents
\item $\mathcal{T}: \mathcal{S} \times \mathcal{A} \rightarrow \mathcal{S}$ is the transition probability distribution $p(s_{t+1}|s_t, a_t)$
\item $\mathcal{R} \subset \mathbb{R}$ is the reward function that determines the goal of the agent
\item $\gamma \in [0,1]$ is the discount factor for cumulative reward computation
\end{itemize}

The learning model and decision-maker is the agent, and the remaining elements outside the agent comprise the environment. The experience of an agent is a sequence of interactions of discrete time steps $t=0,1,2,3,...$.
When an agent interacts with its environment, it observes a state $s_t$ at time step $t$ emitted by the environment, decides the action $a_t$ based on the observation $s_t$ and responds to the environment. Given the state of the last time step $s_t$ and the action $a_t$, the environment transitions to the state of the next time step $s_{t+1}$ and sends an immediate reward $r_{t+1}$ back to the agent. Without further specification,  $s_t$ contains sufficient information for the agent to decide the best action. The agent learns by collecting new experience data and optimizing a policy $\pi$ for action selection.

We consider the finite episodic MDP where the state space and action space are finite. An episodic MDP has finite length of sequence. A finite MDP has finite state space and action space. This is a realistic setting for generation tasks. For example, in de-novo molecular design, the action space is the set of all sub-part molecule embeddings, which includes atoms and bounds between atoms, and the state space is concatenation of actions, forming a string of aotms and bounds. Each molecule can be represented as a finite character sequence, therefore it is episodic. The atom space and the space of bounds between atoms is finite. Another example is image generation. The common approach is generate RGB pixels that have discrete and finite spaces. Given the limited size of an image, the state space and action space of image generation is finite as well.
In an episodic MDP, the environment resets itself after transitioning $T$ steps. The $T$-step sequence is called an episode. 
The goal of an agent is to achieve the highest cumulative rewards from the environment in an episode. At each time step $t$, the agent should select an action that achieves maximum of the cumulative rewards. The agent maps a state to an action by a deterministic policy or a probability distribution $\pi(a_t|s_t)$. Note that the cumulative reward is termed as a return at time step $t$, $R_t=\sum_{k=0}^{T-1-t} \gamma^{k} r_{k+t+1}$, which means that the agent should consider future expected rewards instead of the reward in the current step. $\gamma$ is the discount factor that decreases the impact of future rewards at an exponential rate. Then the objective of an agent is formalized as
\begin{equation}
\label{eq:rl-objective-pi}
    \pi^{*} = \operatorname{argmax}_{\pi} \mathbb{E}[R|\pi]
\end{equation}

One critical assumption of an MDP compared to unconstrained RL tasks is the Markov Property, where 
$$
p(s_{t+1}, r_{t+1}|s_0, a_0, s_1, a_1, ..., s_t, a_t) = p(s_{t+1}, r_{t+1}|s_t, a_t).
$$
It guarantees that state $s_t$ and action $a_t$ determine the next state $s_{t+1}$ and $r_{t+1}$. This property is useful in the sense that the previous state step provides sufficient information, laying the foundation for the efficiency of the algorithms. 

We select research branches that are related to the generation literature. We go through  classic models in the RL research, model-free RL and model-based RL.

\subsubsection{Model-free Methods}
\label{sec:background-model-free}

We first describe  Model-free RL. There are two main approaches: models based on value functions and models based on policy search. Value functions are variants of the objective in Equation \ref{eq:rl-objective-pi}. This objective is an expectation over a sequence of rewards while the interaction unfolds step by step. At time step $t$, the agent's policy is to optimize the objective in the current time step $R_t$. Under a policy $\pi$, the expectation of $R_t$ is the value function
$ V^{\pi}(s_t) = \expectation_{\pi}[R_t|s_t] $ and 
$ Q^{\pi}(s_t, a_t) = \expectation_{\pi}[R_t|s_t, \pi(s_t)]
$.

The value functions have particular recursive properties: 
    $ V^{\pi}(s_t) = \expectation_{s_{t+1}}[r_{t+1} + \gamma V^{\pi}(s_{t+1}) ]$ and
    $Q^{\pi}(s_t, a_t) = \expectation_{s_{t+1}}[ r_{t+1} + \gamma Q^{\pi}(s_{t+1}, \pi(s_{t+1}))]$
which are called the Bellman Equation \cite{bellman1954theory} which makes it possible to train a model through one-step modeling. Bellman Equation could estimate the value at each time step. Compared to methods that estimates the value at the end of each episode, it could decreases the variance of the estimation and make it easier for the model to learn which action has the highest expected return.
Bellman Equation leads to the classic Q-learning that has the following learning rules:
\begin{equation}
\begin{aligned}
    Q^{\pi}(s_t, a_t) & = Q^{\pi}(s_t, a_t) + \\
    & \alpha [ r_{t+1} + \gamma \max_{a_{t+1} \in \mathcal{A}} Q^{\pi}(s_{t+1}, a_{t+1}) - Q^{\pi}(s_t, a_t)]
\end{aligned}
\end{equation}

Classic model-free RL algorithms are listed in the Table \ref{tab:model-free-rl-algo}.

\begin{center}
\begin{table} 
\begin{tabular}{cc}
    \toprule
    RL Algorithms & Related Works \\
    \hline
    Value-based & Q-learning\cite{sutton2018reinforcement}, DQN \cite{mnih2015human}, Soft Q-learning \cite{haarnoja2017reinforcement} \\
    Policy-based & \makecell{Policy Gradient/REINFORCE \cite{williams1992simple}, 
    \\ Actor-Critic \cite{sutton2018reinforcement}, TRPO \cite{schulman2015trust}, PPO \cite{schulman2017proximal}, A3C \cite{mnih2016asynchronous}}\\
    Hybrid Methods & \makecell{DDPG \cite{lillicrap2015continuous}, SAC \cite{haarnoja2018soft},} \\
    \bottomrule
\end{tabular}
\caption{Model-free RL algorithms.}
\label{tab:model-free-rl-algo}
\vspace*{-5mm}
\end{table}
\end{center}

\textbf{DQN} \cite{mnih2015human} extends Q-learning with the neural approximation of Q values. It devises the experience replay method where an interaction history is collected, stored, and used to retrain the parameters of the Q function. This technique smooths the distribution of transitions, which can stablize the training process by randomly sampling from memory instead of correlated episodes sampled from recent states of environments.

Different from value-based methods, \textbf{Policy gradient} or REINFORCE\cite{williams1992simple} is a method that directly maximizes the objective by computing the gradient of the policy:
\begin{equation}
\begin{aligned}
\nabla_{\theta} \mathbb{E}[R|\pi] = \mathbb{E}_{\tau \sim \pi_{\theta}}[ r(\tau) \cdot \nabla_{\theta} \log \pi_{\theta}(\tau) ] \\
\theta_{t+1} = \theta_{t} + \alpha r(\tau) \cdot \nabla_{\theta} \log \pi_{\theta}(\tau) \\
r(\tau) \cdot \nabla_{\theta} \log \pi_{\theta}(\tau) = \sum_{t=1}^T R_t \log \pi_{\theta}(a_t| s_t)
\end{aligned}
\end{equation}
where $\tau$ is defined as a trajectory variable that includes states and actions.

\textbf{REINFORCE with baseline} \cite{williams1992simple,sutton2018reinforcement} is defined to subtract the value term with a baseline term. The baseline term can be any function, even a random term that takes the state as an input:
\begin{equation}
\begin{aligned}
\theta_{i+1} = \theta_{i} + \alpha(R_t - b(s_t)) \nabla_{\theta} \log \pi_{\theta}(s_t) \label{eq:reinforce-baseline}
\end{aligned}
\end{equation}
where $i$ represents the iteration of model parameter update, $R_t$ is the expected return term. $b(s_t)$ is the baseline. This results in an unbiased estimation of the policy with a reduced variance\cite{sutton2018reinforcement}. A natural choice for the baseline is the state value function $V^{\pi}(s_t)$ that can capture the value fluctuations of different states.

\textbf{Actor Critic} methods \cite{sutton2018reinforcement} have a similar form as variance-reduced REINFORCE algorithm does, which is a branch of policy-based methods. The key difference is that the value function takes part in the value term of Equation \ref{eq:reinforce-baseline}, which can increase the bias of estimation and accelerate learning by decreasing the variance, which is formulated by
\begin{equation}
\begin{aligned}
\theta_{t+1} = \theta_{t} + \alpha(r_{t+1} + \gamma \hat{V}^{\pi}(s_{t+1}) - \hat{V}^{\pi}(s_{t}) ) \nabla_{\theta} \log \pi_{\theta}(s_t) \label{eq:actor-critic}
\end{aligned}
\end{equation}
where the value term uses both a policy model $\pi_{\theta}(s)$ and an estimated value function $\hat{V}^{\pi}(s)$ for bootstrapping. They can benefit from both value-based methods and policy-based method. On the one hand, the selection of actions for value functions requires the maximum values of all actions. When there is an infinite number of actions, it can be intractable to compute all the values. Policy methods can remedy this by action selection  instead of value evaluation in value-based methods.  On the other hand, policy models suffer from the high variance that can be alleviated by bootstrapping to accelerate learning. 

In actor critic method, another major innovation is the branch of Trust Region Policy Optimization (\textbf{TRPO}) \cite{schulman2015trust} and Proximal Policy Optimization \cite{schulman2017proximal}. TRPO \cite{schulman2015trust} introduces a trust region for the policy to update itself, in which the policy improvement is monotonically guaranteed theoretically. The trust region algorithm is used to compute the gradient of the following constrained optimization problem:
\begin{equation}
\begin{aligned}
\operatorname*{max}_{\theta} &\expectation_{s \sim p_{\theta_{old}}, a \sim \pi_{\theta_{old}}} \left[ \frac{\pi_{\theta}(a|s)}{\pi_{\theta_{old}}(a|s)} Q_{\theta_{old}}(s,a) \right] \\
s.t. &\expectation_{s \sim p_{\theta_{old}}} \left[ D_{KL}(\pi_{\theta_{old}}(\cdot|s) \;\|\; \pi_{\theta}(\cdot|s) \right] \leq \delta \label{eq:trpo}
\end{aligned}
\end{equation}
where the objective function is the actor critic method that uses the trajectories of an old policy $\pi_{\theta_{old}}$ and computes the expected return with an old Q function $Q_{\theta_{old}}$. The objective is constrained by an expectation of KL divergence between the old policy and the new policy. This problem is estimated by a quadratic approximation solved by the conjugate gradient method. \textbf{PPO}\cite{schulman2017proximal} uses a clipped surrogate objective to form a lower bound of the objective of TRPO. Replacing the complicated quadratic approximation, PPO utilizes only first-order optimization to solve a constrained problem:
\begin{equation}
\begin{aligned}
\operatorname*{max}_{\theta} &\expectation_{s \sim p_{\theta_{old}}, a \sim \pi_{\theta_{old}}} \left[ \operatorname{min}(f_{factored\_policy}(s, a) Q_{\theta_{old}}(s,a), \right.  \\
& \left. \operatorname{clip}(f_{factored\_policy}(s,a), 1 - \epsilon, 1 + \epsilon) Q_{\theta_{old}}(s,a)  ) \right] \\
 \label{eq:ppo}
\end{aligned}
\end{equation}
where $f_{factored\_policy}(s,a)=\frac{\pi_{\theta}(a|s)}{\pi_{\theta_{old}}(a|s)}$. This formulation substitutes the KL divergence constraint as a clip operation to control the divergence between $\pi_{\theta_{old}}$ and $\pi{\theta}$ that enables the algorithm to replace the complex optimization process.

Another line of research in actor-critic models is to accelerate training through parallel computation. The asynchronous advantage is actor-critic (\textbf{A3C}) \cite{mnih2016asynchronous} is the representative example. The policy and value parameters are updated asynchronously. The model does not require a replay buffer because it runs multiple agents in parallel with different exploration policies that can stabilize training. It achieves good performance increases compared to DQN, Sarsa, and n-step Q.


Apart from value-based and policy-based methods, hybrid methods tend to integrate the advantages of both methods to strike a balance.
\textbf{DPG and DDPG} Silver et al. \cite{silver2014deterministic} proposed the Deterministic Policy Gradient (DPG) algorithms. It allows a more efficient estimation of policy compared to policy gradient. DPG does not estimate the value $Q^{\pi}$ by the Monte Carlo method, it uses an actor-critic method to estimate the Q value and compute the gradient of actions directly by
\begin{equation}
\begin{aligned}
    \nabla J_{\theta} = \expectation_{s \sim d^{\pi}} \bigg[ \nabla_{\theta}\pi_{\theta}(s) \;\; \nabla_{a} Q^{\pi}(s,a)\bigg\rvert_{a=\pi_{\theta}(s)} \bigg]
\end{aligned}
\end{equation}
The difference between Monte Carlo based policy gradient and actor-critic is that the Monte Carlo method requires to estimate the expected reward of a state by averaging all returns of visits to a state. This requires to iterate from $T$ to $0$, collect returns, and compute averages. Actor-critic is more easy to compute by the Bellman Equation.
Lillicrap et al. \cite{lillicrap2015continuous} extend  DPG to Deep DPG (DDPG), which combines training techniques from DQN \cite{mnih2015human}. 
DDPG incorporates experience replay and a soft target that updates the model parameters with a control variable to slow down the parameter changes. 

Soft Q-learning \cite{haarnoja2017reinforcement} derives an energy-based policy for continuous states and actions by modeling the policy distribution as a Boltzmann distribution. 
The reward function is modified to
$
R_t=\sum_{k=0}^{T-1-k} \gamma^{k} ( r_{k+t+1} + \mathcal{H}(\pi(\cdot|s_{k+t+1})) )
$
Soft actor-critic (SAC) \cite{haarnoja2018soft} integrates the Soft Q-learning into actor-critic methods and optimizes towards the direction of rewards plus entropy of policy distributions.


\subsubsection{Model-based Methods}

In RL, models come from prior knowledge or learning. For example, for an agent playing Go, the rules of Go are fixed, thus it's possible to get a perfect environment model by programming. Influential works like AlphaGo \cite{silver2016mastering} constructs a Monte-Carlo Tree for forward prediction. Sometimes it's not feasible to get a perfect model. In AlphaGo \cite{silver2016mastering},  policy learning is integrated with Monte-Carlo Tree search for the Go game. Generally, Monte-Carlo Tree search creates a tree whose nodes represent states in RL. The expansion of the tree is exploring actions and new states. Decision making is based on values on nodes. AlphaGo adds a policy network and a value network to the tree. During inference, nodes are explored and values are obtained along the tree structure. The parameters of the policy network and the value network are updated in training by policy gradient and mean squared error respectively. Actions are selected with three factors in consideration, a Q-value, a probability and a number of traversals. The Q-value is computed by mixing a state value and a reward collected from random fast rollouts. AlphaZero \cite{silver2017mastering} focuses on self-play to learn a policy from scratch using an adapted version of AlphaGo models.

Dyna Q-learning \cite{sutton1990integrated} employs the model with domain knowledge as a predictor and data augmentor for the model-free policy, combining trial-and-error, a domain knowledge model, planning, and reactive execution into one algorithm.  The dynamics model generates pseudo-experiences that are incorporated into policy training.

\subsubsection{Comparison Between Reinforcement Methods}
Given the fact that there are various types of RL methods, it could be difficult to select from them for a specific generation task. Generally, model-based RL requires to use a world model or learn one, in contrast to model-free methods. DQN \cite{mnih2015human} is more suitable for discrete actions and policy gradients-based methods and actor-critic methods could handle continuous space. TRPO \cite{schulman2015trust} and PPO \cite{schulman2017proximal} enhance the stability for learning. DDPG \cite{lillicrap2015continuous} learns a deterministic policy.
We suggest that researchers should start with basic algorithms like DQN or policy gradient, then investigate towards more complex algorithm training method. From the survey we observe that more works employ policy gradients than DQN. This could originate the characteristics of policy gradients and DQN. 
For example, PPO \cite{schulman2017proximal} is popular in fine-tuning a large language model, while a recent work \cite{ahmadian2024back} shows that the performance of PPO \cite{schulman2017proximal} could be matched by well RLOO, a well designed REINFORCE algorithm \cite{williams1992simple} by investigating the components that works for large language model pre-training. 
This work argues that the PPO aims to improve stability of training for environments exploration with high variance in RL tasks, while it is not suitable for RLHF fine-tuning which has a relatively stable initial distribution. It replaces PPO with REINFORCEMENT algorithm without partial sequence reward learning which accelerates learning and achieves better performance.

\subsection{Framing Generation tasks as Reinforcement Learning Problem}
\begin{figure*}[t]
\centering
\subfloat[Reinforcement learning agent as generator. The action is integrated into the observation in the next step.]{
  \includegraphics[width=0.45\linewidth]{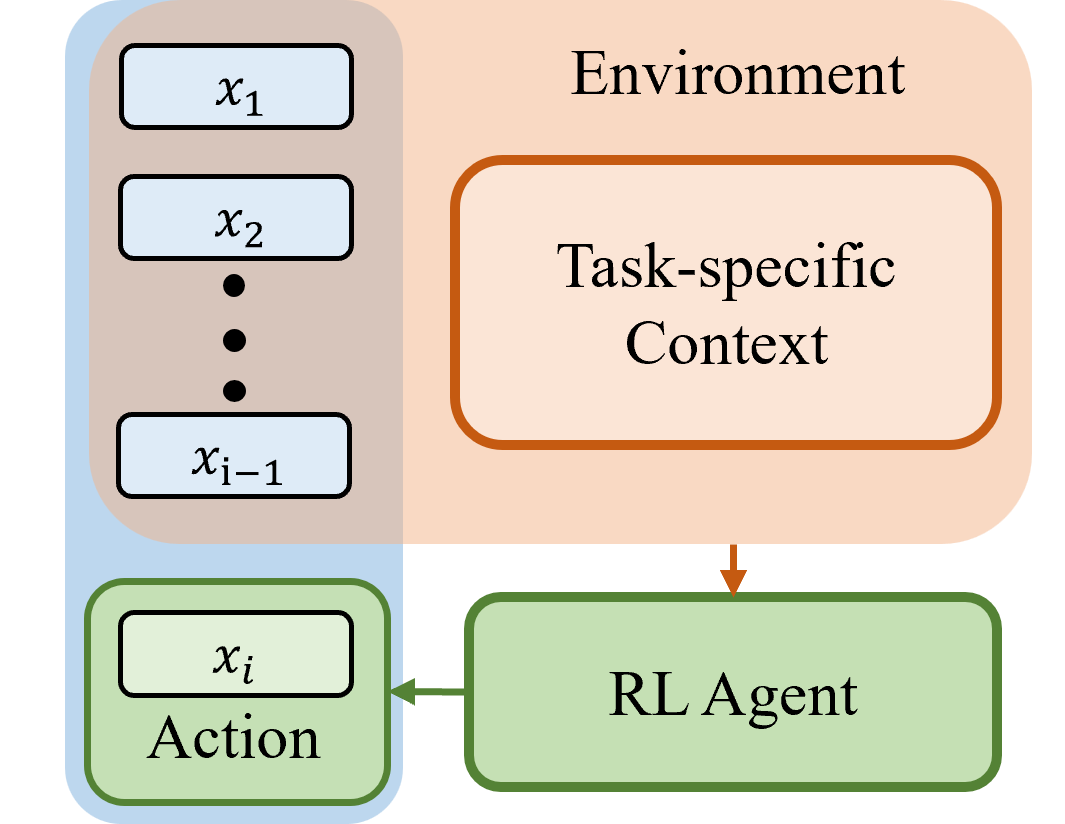}
  \label{fig:rl-as-gen}
}%
\subfloat[Non-differentiable setting for generation models. The blue boxes refer to steps that might cause non-differentiable condition. Dashed lines represent the block of gradient and the green line show that RL uses signal to train the generator.]{
  \includegraphics[width=0.4\linewidth]{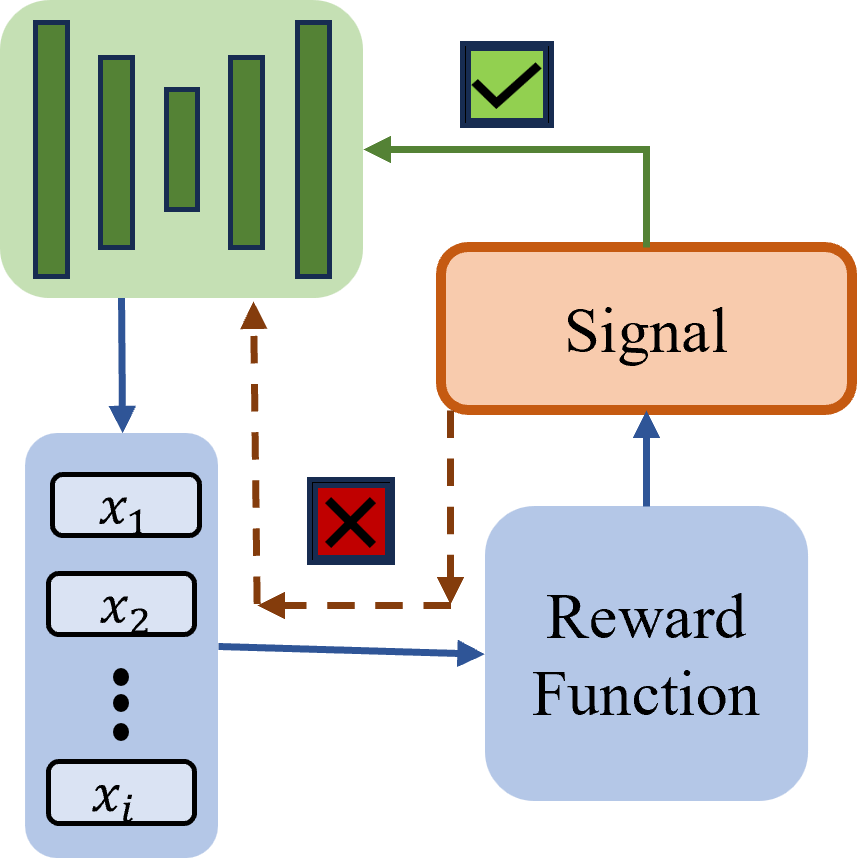}
  \label{fig:non-differentiable}
}
\caption{(a) Framing generation tasks as reinforcement learning framework; (b) The reward computation is non-differentiable.}
\end{figure*}

We use Figure \ref{fig:rl-as-gen} to show how a generator could be framed as a reinforcement learning agent in applications.
Reinforcement Learning problem is typically defined as a MDP as shown in Section \ref{sec:background-rl} which is typically a sequential decision maker. therefore, we use the agent to generate a sequence $x_1, x_2, x_3, ..., x_n$. At time step $t$, the previously generated actions $x_1, .., x_{t-1}$ and the task-specific context form the data of the environment. For example, when the application is a visual question answering, $x_1, ..., x_{t-1}$ are answer tokens, task specific context includes the image and the token sequence of the question.

\section{Benefits of RL-based Generative Models}\label{sec:benefits}


\subsection{Solving the non-differentiable learning problems}

One major use of reinforcement learning is that it can propagate gradients through non-differential modules. This extends the capability of neural networks because it allows the model to be trained when discrete modules exist in the computation pipeline. This characteristic is supplementary to supervised learning and unsupervised learning objectives, which both require a differentiable training pipeline. In this section, we introduce two classes of non-differentiable problems.

\subsubsection{The generated variable is non-differentiable}
\label{sec:non-diff-variable}

Discrete values are prevalent in various generative applications such as computer vision, neural language processing, and molecule generation.  In language applications and molecular design, elements of text and molecules are tokenized and embedded into high-dimensional space in order to capture a better representation. The tokens are discrete values or one-hot vectors. In computer vision, a common format of images is the RGB format which comprises discrete values of three color channels. Although it is feasible and easy to normalize the discrete values that are fed into a continuous generative model, transforming discrete values into continuous values leads to adverse effects such as weaker robustness \cite{mao2022discrete}.

Reinforcement learning is a suitable tool for such problems. The policy gradient method is a widely adopted approach in the field of machine learning. The formulation lacks any explicit constraint governing the relationship between the gradient $\nabla_{\theta} \pi_{\theta}(s)$ and the reward $r(\tau)$. The utilization of a reward signal for policy training has been explored in previous works such as\cite{devon2017boundary,fan2020recurrent,nguyen2019infocnf}, circuvmenting the differentiable requirement of supervised learning. For example, BGAN \cite{devon2017boundary} aims to tackle the inherent limitation of GANs in handling discrete data due to the absence of differentiable conditions. This is achieved by establishing a connection between policy gradient and the GAN objective. 
The data distribution is optimized by Monte-Carlo estimation,thereby mitigating variance in the policy gradient.
In language modeling, \cite{fan2020recurrent} borrows the SeqGAN \cite{yu2017seqgan} model into a visual dialog system. The generative agent outputs non-differentiable word sequences. Hence, policy gradient is leveraged as a means of facilitating knowledge transfer.

Moreover, the reinforcement learning agent can control the training instead of being the generator. In this context, the control values can be discrete. For instance, in InfoNCF \cite{nguyen2019infocnf}, the policy gradient is adopted to optimize the number of evaluation functions for normalizing flows in the latent space. 

\subsubsection{The training objective is non-differentiable}

Apart from the generative variable, the training objective can be non-differentiable. Take policy gradient as an example, it allows directly injecting a non-differentiable objective as reward without further constraints. Widespread evaluation metrics for machine translation and text summarization are good examples. For example, MIXER \cite{ranzato2015sequence} proposes to address the exposure bias between the training and testing phase in sequence generation. Consequently, it uses test metrics, BLEU\cite{papineni2002bleu} and ROUGE\cite{lin2004rouge}, to directly optimize the model. This section overlaps with section \ref{sec:introducing-new_training_signal} and we leave other related discussions later.

\subsection{Introducing new training signal}
\label{sec:introducing-new_training_signal}
\begin{figure*}[t]
\centering
\includegraphics[width=0.8\linewidth]{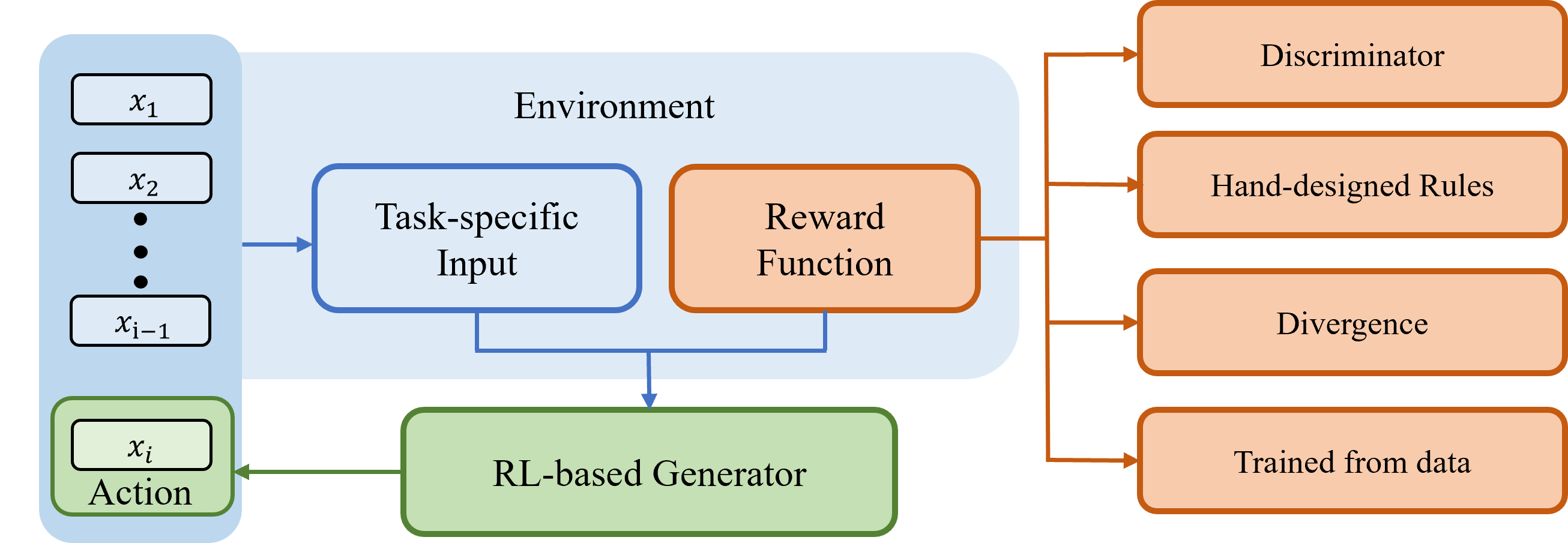}
\caption{RL can introduce new signals by flexible reward functions}
\label{fig:introduce-new-signal}
\end{figure*}

As described in the last section, reinforcement learning methods can be employed for non-differentiable problems that exist in generation applications. In Section \ref{sec:non-diff-variable}, policy gradient has no requirements between reward and policy, in contrast to supervised learning or unsupervised learning, allowing more flexibility. This flexibility exists in most reinforcement learning methods in Section \ref{sec:background-rl}. Thus it is straightforward to design useful reward functions as additional training objectives, which can incorporate various training signals into the generation process, making RL an influential approach in generation model and application. We demonstrate four major approaches in this section, as shown in Figure \ref{fig:introduce-new-signal}.

\begin{center}
\begin{table} 
\caption{Methods to introduce new training signal}
\begin{tabular}{cc}
    \toprule
    What signals could be incorporated by RL & Related Works \\
    \hline
    Probability from a discriminator & \cite{yu2017seqgan,guimaraes2017objective,wang2020grl,fedus2018maskgan,lin2017adversarial,zhou2020self,scialom2020coldgans, che2017maximum,scialom2021beam,
sarmad2019rl,wu2021textgail}, \\
    Task-specific Hand-designed Rules & \cite{ranzato2015sequence, wan2018improving, li2016deep, mo2018personalizing,
zhao2016towards}\\
    Distribution Discrepancy &  \cite{pyatkin2022reinforced,
fan2023optimizing,
jaques2017sequence,ziegler2019fine,jaques2019way,pang2020text,
norouzi2016reward,ke2019araml,lamprier2022generative}\\
    Train a reward model with human labelled data & \cite{shi2018toward, stiennon2020learning,shi2018toward,ouyang2022training,bai2022training,gao2020preference,kreutzer2021offline,
bai2022constitutional,zhu2023principled}\\
    \bottomrule
\end{tabular}
\label{tab:introduce-new-training-signal}
\end{table}
\end{center}

\subsubsection{Reward by Discriminator}
From the standpoint of training signal, the discriminator component within the Generative Adversarial Network (GAN) architecture fulfills a role akin to that of a reward in the context of reinforcement learning.
SeqGAN \cite{yu2017seqgan} first proposes to exploit this similarity by introducing a GAN to generate the sequence tokens. The reward signal is derived from the output probability, which serves as a discriminative measure between real and generated samples.
Subsequent studies have expanded upon or altered the aforementioned framework through the utilization of a meticulously crafted discriminator \cite{guimaraes2017objective}, the development of novel reward formulations \cite{wang2020grl}, the implementation of actor-critic techniques \cite{fedus2018maskgan}, the incorporation of rank formulations \cite{lin2017adversarial}, the utilization of multiple discriminators \cite{zhou2020self}, and other related modifications.

Objective-reinforced GAN \cite{guimaraes2017objective} extends  SeqGAN with domain-specific objectives to the reward to adapt the generated samples towards the domain-specific direction.
Adversarial rewards are also employed in GRL \cite{wang2020grl} that takes the difference between two probabilities $\expectation_{x \sim p_{\mathit{data}}} D(x) - \expectation_{x \sim \pi_{\theta}} D(x)$.
MaskGAN \cite{fedus2018maskgan} replaces the baseline in REINFORCE by a critic, forming an actor-critic algorithm for text generation instead of a policy gradient in SeqGAN. It uses an in-filling task to train the agent and uses the probability of real words in the discriminator as the reward.
SAL \cite{zhou2020self} changes the reward function with comparison discriminators. The discriminators take a pair of samples $(x_1, x_2)$ as input, which is collected from the current generated sample and previous ones. Three types of discriminators $D^{(>)}, D^{(<)}$, and $D^{(\approx)}$ are defined to describe the relationship between the quality samples. Using coefficients for a better balance between exploration and exploitation has been taken into consideration.
RankGAN \cite{lin2017adversarial} follows the combination of GAN and RL, which substitutes the discriminator with a ranker and the reward is provided by a rank score estimated by the ranker given a sentence and a reference set and a comparison set. The ranker is trained to work as a discriminator by increasing the score of sentences from the dataset and decreasing the score from the generator. 
Li et al. \cite{li2017adversarial} introduce adversarial learning into the RL-based dialogue generation framework, where the reward is defined as the score produced by a discriminator on the dataset consisting of a human-generated response or auto-generated response. This method uses the RL framework to guide the generated texts by human text distribution. To prevent deteriorating the model quality, a teacher-forcing method that includes human supervision in discrimination training is adopted.
ColdGAN \cite{scialom2020coldgans} explores the exposure bias problem on the GAN-based model in the text generation tasks. It analyzes the impact of randomness on discriminators and finds that bad discriminators might mislead the generators to low-quality areas of the parameter space. Therefore, it integrates the constraints of old policies as an importance sampling strategy to balance the impact of the discriminator. It also proposes a policy merge method for a cautious generative process.
MaliGAN \cite{che2017maximum} combines maximum likelihood with gradient descent. It hypothesizes that the discriminator is easy to learn and can get optimal results. Under this assumption, the following equation holds,
$$
\mathbb{E}_{p_\mathit{d}} [\log p_{\theta}(x)] = \frac{1}{Z(\theta^{'})} \mathbb{E}_{p^{'}} [r(D) \log p_{\theta}(x)]
$$
where $Z(\theta^{'}) = \mathbb{E}_{p^{'}} [r(D)] = 1$. When the discriminator is not sufficiently trained, the gradient of KL divergence almost surely is larger than 0. Based on this, an importance sampling is designed with baseline as reward $\frac{r_D(x)}{\sum r_D(x)} - b$. MCTS is introduced to cope with high variance.
SelfGAN \cite{scialom2021beam} incorporates MCTS to cooperative decoding to tackle the instabilities of training the discrete sequential decision-making problems such as summarization and question answering. Cooperative decoding means that discriminators are employed in the decoding phase to rerank the generated sequences as value functions. As in MCTS, action is selected based on a Q-value, policy probability, and counts of traversals. Therefore, the discriminator is integrated into the MCTS decoding procedure to improve the performance.
RL-GAN-Net \cite{sarmad2019rl} proposes to use an RL agent to generate latent code for the GAN model on point cloud generation. The reward comprises discriminator loss, point cloud reconstruction loss, and  Chamfer distance-based loss.

TextGAIL \cite{wu2021textgail} incorporates a contrastive discriminator where the input includes the previous sequence as well as real data or generated data. The discriminator is required to evaluate the relative realness between sequences. It also employs a PPO \cite{schulman2017proximal} to decrease the variance of the RL training.

\subsubsection{Reward by Hand-designed rules}
Novel metrics or heuristic functions are intended to provide incentives for training, and as far as we can tell, the majority of RL algorithm implementations fall within this branch.

Hand-designed rules could lead to non-differentiable objectives \cite{ranzato2015sequence,rennie2017self,cai2020tag,dognin2021regen}. 
Like MIXER\cite{ranzato2015sequence}, SCST \cite{rennie2017self} defines the reward by the performance of the current model under the inference algorithm, including CIDEr\cite{vedantam2015cider}, BLEU4\cite{papineni2002bleu}, ROUGEL\cite{lin2004rouge}, and METEOR\cite{banerjee2005meteor}.
TAG \cite{cai2020tag} incorporates type auxiliary guiding for code comment generation by reinforcement learning. The model takes BLEU\cite{papineni2002bleu} and ROUGE\cite{lin2004rouge} as rewards.
ReGen \cite{dognin2021regen} proposes to use reinforcement learning for non-differentiable evaluations like BLEU\cite{papineni2002bleu}, METEOR\cite{banerjee2005meteor}, and chrF++\cite{popovic2017chrf++} to guide  text generation.

One line of research is incorporating testing-time metrics as a reward. This line overlaps the branch of non-differentiable objective when the evaluation metric is non-differentiable, such as ROUGE \cite{narayan2018ranking,chen2018fast,akyurek2023rl4f,ranzato2015sequence} and BLEU \cite{ranzato2015sequence,wan2018improving}.
The utilization of testing-time metrics also enhances models to combat the discrepancy between the training objective and testing objective.
MIXER \cite{ranzato2015sequence} proposes to address this problem in sequence generation by reinforcement learning. It applies REINFORCE to train an agent whose actions are next words given the current time step context. The reward is the test metric for neural language processing models.
Wan et al. \cite{wan2018improving} introduce an actor-critic algorithm to code summarization by leveraging the exploration feature in reinforcement learning.

Apart from testing-time metrics, task-specific reward functions are also popular.
Li et al. \cite{li2016deep} propose to use a policy gradient with three considerations in dialogue generation, ease of answering, information flow, and semantic coherence. The ease of answering is measured by the negative log-likelihood
$- \frac{1}{N_{\mathcal{S}}} \sum_{b} \log p(b|a)$ where $b$ is a human constructed list of dull responses that contains sentences like ``I don’t know what you are talking about'', and $a$ is the generated sentence.
 The information flow is constructed by penalizing similarity between two consecutive turns of the same agent. The semantic coherence is measured by the mutual information between this turn action and actions from previous turns.
PETAL \cite{mo2018personalizing} manages dialog systems by reinforcement learning. It studies dialog systems in real-world coffee order systems. The reward contains multiple items, representing personal reward and general reward. The personal reward reveals the interaction between the agent and the user, such as accepting or rejecting  the agent's suggestion. The global reward includes motivation for getting the user's information, payment, and shortening the dialog. In modeling, it breaks down the value function into two parts, one standing for general value, and one for personal preference. This achieves a better transfer effect when a model is tested on new users.
Zhao et al. \cite{zhao2016towards} employs RL to interact with a database in a dialog state tracking and management task, which is beyond the capability of supervised learning on this task. The agent is not only asked to respond to users but also query from a database to better manage the dialog. Databases are used to generate synthetic data, which are combined with real data for value function learning and policy learning, like in Dyna Q-learning \cite{sutton1990integrated}. 

\subsubsection{Reward by Distribution Discrepancy}
\label{sec:benefits-new-signal}
Distribution discrepancy can be a useful signal to be integrated into rewards. Maximizing the divergence leads to more informative generative data\cite{pyatkin2022reinforced}, minimizing the divergence between generated distribution and some distribution could regularize the generation \cite{fan2023optimizing,jaques2017sequence,ziegler2019fine,jaques2019way,pang2020text}.
CLARIFYDELPHI \cite{pyatkin2022reinforced} uses RL to generate clarification questions to elicit moral judgment of models. The question is generated by a PPO network whose reward is the divergence between two different moral judgments of the questions. An answer simulation framework is designed to get the divergence between different answers.

Motivated by the policy gradient, Fan and Lee \cite{fan2023optimizing} utilizes distribution discrepancies to optimize DDPM sampling with shortcut fine-tuning. The goal is to use gradient-like optimization algorithm to explore alternative paths to discover more efficient paths and boost speed of sampling by replacing the backward process of DDPM during fine-tuning. The policy is initialized with a trained DDPM generator, which is then guided by a generalized critic function that serves as a measure of discrepancy between the distribution of generated data and real data, namely a generalized divergence. The gradient of a DDPM sampler is formulated to a REINFORCE with baseline mentioned in Section \ref{sec:background-model-free}, The reward is computed based on the generalized critic function that incorporates 1-Lipschitz functions as regularization. The critic function is instantiated as a neural network which is trained to minimize the reward (maximize the discrepancy between distributions) and the policy is trained to maximize the reward (minimize the discrepancy).

When applying reinforcement learning algorithms to generation models and applications, a line of research combines reinforcement learning with supervised learning to guarantee that the model is adjusted by the reward signals but does not drift away from the supervised training objective to prevent the model from generating highly rewarded but unrealistic results. The divergence between a generated distribution and the distribution defined in the dataset is explored\cite{jaques2017sequence,ziegler2019fine,jaques2019way,pang2020text}. KL-control is a technique for non-Markovian systems to minimize deviation from a prior policy\cite{jaques2017sequence,todorov2006linearly}.
Sequence Tutor \cite{jaques2017sequence} integrates a KL control method in order to maintain a policy generation that remains in proximity to the pre-trained language model. The reward function incorporates previous knowledge derived from a pre-trained recurrent neural network (RNN) model..
Ziegler et al. \cite{ziegler2019fine} incorporate human preference learning into pre-trained language models. It involves KL-control for coherence and topicality.
Jaques et al. \cite{jaques2019way} inject KL-control into discrete Q learning to impose an entropy regularization.
GOLD \cite{pang2020text} aims to address two problems in the MLE training paradigm in text generation: diverse but low-quality samples and exposure bias. Unlike the studies mentioned above, it adopts an offline reinforcement learning algorithm. It uses a weighted policy gradient where the weights come from the training set policy because, in offline reinforcement learning, the agent cannot sample trajectories to estimate the weight. The weight reveals the conservative method: keep the actions on the test dataset similar to the training dataset. For reward, it uses training trajectories to approximate the probability distribution of human preference. It combines three kinds of rewards that use a one-zero reward, the product of MLE probability, and the sum of MLE probability.

\subsubsection{Reward by data-driven model}
With the flexibility of the reward function, it is also a good way to incorporate models learned from the reward function. It is feasible to incorporate various guidance into reinforcement learning by training a reward model. This branch expresses a similar idea as Inverse Reinforcement Learning (IRL), which can be integrated in generative models in two ways.

One is directly embedding previously defined rewards such as BLEU into a model. Shi et al. \cite{shi2018toward} conduct experiments in text generation and empirically prove its effectiveness. They adopt the maximum entropy IRL to model an approximated reward function. The training process of the reward model increases the rewards of real texts and decreases the rewards of texts that are sampled from the approximated distribution by a generator with importance sampling. Furthermore, an entropy term is added to the reward function for the agent to prevent premature mode collapse and increase the diversity of generated texts.

The other way is to learn a model for human preference \cite{gao2019reward,nguyen2017reinforcement}. This path finally leads to the emergence of Reinforcement Learning Human Feedback (RLHF) \cite{stiennon2020learning,shi2018toward,ouyang2022training,bai2022training,gao2020preference,kreutzer2021offline}, which is integrated into the large language model research and harvest powerful models like ChatGPT \cite{openai2022chatgpt}.
RELIS \cite{gao2019reward} proposes to learn a reward function from learning to rank objectives on a document summarization task. Human preferences of two summaries in the form of ranking are collected and train the reward model by three types of loss: cross entropy, marginal ranking, and an improved marginal ranking. It is theoretically proved that the agent converges to a near-optimal solution.
Nguyen et al. \cite{nguyen2017reinforcement} study the simulated human feedback in the form of ratings in neural machine translation. They are motivated by the fact that human feedback is not perfect. For example, expert ratings cannot perfectly match the goal. There are also granularity, variance, and skewness problems in the collected ratings. Therefore, they propose to simulate human feedback and address the problems mentioned above. They map the feedback to binned values, use a linear approximation to deal with large variances in middle ratings and employ a skew perturbation for harsh and motivational scores.
The
Open AI Reflection team \cite{stiennon2020learning} uses human preference to guide language models for summarization tasks. The training consists of three steps. First, trajectories from a trained policy with various baselines are collected and these trajectories are evaluated by humans to rank the best one. Then, they construct a model to learn the rewards that indicate whether the output is better. Last, they optimize a policy given the reward model. It is similar to the step in \cite{shi2018toward} while it combines datasets from human preferences, which dramatically outperforms existing methods at that time.
The reward function is trained by the following function,
$$
\operatorname{loss}(r_{\theta}) = E_{(x,y_0,y_1,p)}[\log (\sigma(r_{\theta}(x, y_p) - r_{\theta}(x, y_{1-p})))]
$$
where $x$ is the text before summarization, $y$ is summarized text, $r_{\theta}$ is the reward function, $y_p$ is the human preferred text. The loss aims to maximize the distance between two rewards. Additionally, authors use KL-control \cite{jaques2017sequence} to prevent the mode collapse as well as constrain the policy to be conservative, not generating weird texts far from the original supervised pre-trained distribution.
InstructGPT \cite{ouyang2022training} follows the same procedure to fine-tune  GPT-3 from human feedback. Results show that it improves the GPT-3 on truthfulness and generalization, and decreases the toxicity and performance regressions.
Bai et al. \cite{bai2022training} follow the \cite{stiennon2020learning} work to test  RLHF on a helpful and harmless dataset. They use a PPO to train the model and use the same pipeline to learn a preference model and finetune the language model with reinforcement learning. It tests the model in an iterative online mode of training and shows that it improves the performance of the model. It also identifies a roughly linear trend between the preference reward and the square root of the KL divergence between the policy  and its initialization.
APRIL \cite{gao2020preference} combines preference learning with neural TD, an algorithm that replaces the linear approximation in Linear TD \cite{sutton1984temporal} with a neural network. Preference learning employs the  cross-entropy between true preference and the model used to train a reward model. For the limits of human feedback collection, a pair-generation method is proposed to make the process efficient. The pair is generated on the metric of utility gap, diversity, density, and uncertainty. Given a human text $y^{'}$, the utility gap is used to maximize the gap to get high-quality negative samples. Other three metrics aim to make the selected sample diverse, located in a dense part of the distribution, and uncertain. The neural TD is used instead of DQN because the action space is large and the maintenance of Q-value is expensive.
Kreutzer et al. \cite{kreutzer2021offline} discuss the necessity, challenges, and potential solutions of offline reinforcement learning from human feedback. The necessity of offline RL is that online adjustment of parameters is too risky and potentially out of control. The challenges include questionable counterfactual estimation for the lack of explicit exploration and degeneration problems where low-reward actions are still encouraged during training. Also, reliability and learnability are also discussed.

Moreover, the fast development of large language models inspires researchers to use them as a reward function by proper prompting. For example, Constitutional AI \cite{bai2022constitutional} proposes RLAIF that trains a harmless but non-evasive AI assistant that copes with harmful queries via expression of objection to these queries. Self-critiques and automatic revisions from an LLM are exploited to modify the dataset and retrain the LLM on it. Then, an agent is trained by preferences given by humans and models. Human provides helpfulness evaluations while the model provides harmlessness evaluations. The label is generated by an assistant model under the context that prompts contain human set principles as well as a set of few-shot examples. Apart from peakiness, Zhu et al. \cite{zhu2023principled} provide theoretical support for RLHF. They provide a sample complexity for the union problem of RLHF and max-entropy Inverse Reinforcement Learning. They frame the ranking-based reward model as a Plackett-Luce (PL) model or a  Bradley-Terry-Luce (BTL) model, providing suboptimality bound for the reward learning process.

\subsection{Sampling}
\begin{figure*}[t]
\centering
\includegraphics[width=0.85\linewidth]{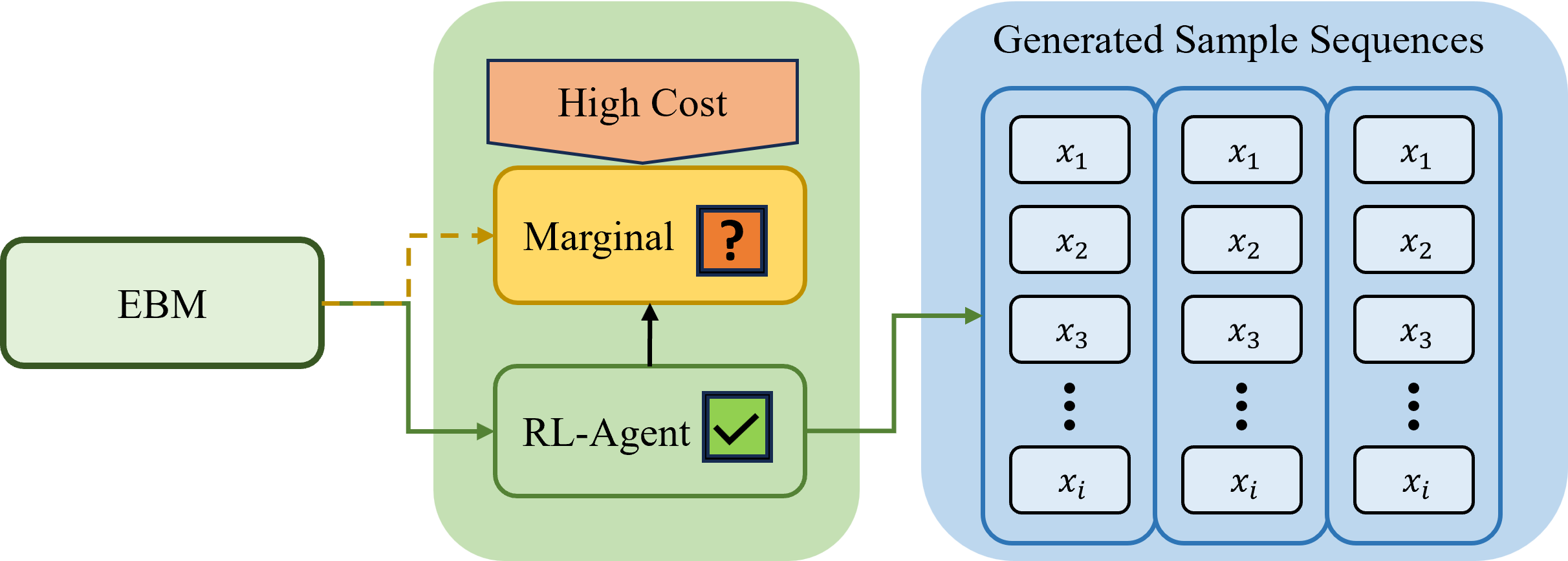}
\caption{RL can work as a sampler for models that are hard to sample such as Energy-based Models. The marginal distribution is difficult to sample for the high cost, RL-based agent provides an alternative way to generate sample sequences. THe dashed line represents the potential high cost blocks the generation.}
\label{fig:rl-as-distributional}
\end{figure*}
Although Energy-based Models, one type of mainstream generative models, enjoy greater expressivity and allow global constraints, they face challenges in producing samples of marginal distribution for the unnormalized distributions in the formulation, which might incur high cost during sampling, as shown in Figure \ref{fig:rl-as-distributional}. Reinforcement learning could be an alternative way to train a sampler for the EBM. A recent line of research \cite{parshakova2019distributional,khalifa2020distributional,korbak2021energy,korbak2022on,go2023aligning} studies the integration of reinforcement learning algorithm to distill knowledge in an EBM into an auto-regressive model by turning distribution matching into the training signal for the RL algorithm. Distributional Policy Gradient \cite{parshakova2019distributional} is a pioneer work that transforms an Energy-Based Model (EBM) training process into a policy gradient algorithm. Employing reinforcement learning (RL) as a pipe to train a sampling generator, \cite{parshakova2019distributional} suggests addressing the distribution learning problem in Global Autoregressive Models (GAM), a subset of Energy-Based models. There are two phases to the training procedure. 
The MLE criteria are used to train the autoregressive factor on the dataset in the first step. 

\begin{table} 
\centering
\caption{Methods in Sampling and NAS}
\begin{tabular}{cc}
    \toprule
    Methods & Related Works \\
    \hline
    Sampling & \cite{parshakova2019distributional,khalifa2020distributional,korbak2021energy,korbak2022on,go2023aligning}, \\
    NAS & \cite{zoph2016neural,pham2018efficient,hsu2018monas,pham2018towards,guo2019irlas,zhong2018practical,tian2020off,chen2020catch,baker2016designing,pang2021rl,chauhan2023dqnas}\\
    \bottomrule
\end{tabular}
\label{tab:sampling-nas}
\end{table}
 
The second stage is the focus of \cite{parshakova2019distributional}, where a policy $\pi$ is used to approximate a desired distribution $p$ via the process of distillation, which facilitates the acquisition of the model parameters. The policy employs a distribution match as a reward and is trained using policy gradient via the process of deductive reasoning based on cross-entropy
\begin{equation}
\begin{aligned}
\nabla_{\theta} \operatorname{CE}(p, \pi_{\theta}) &= - \expectation_{x \sim p(\cdot)} \nabla_{\theta} \log \pi_{\theta}(x) \\
&= - \frac{1}{Z} \expectation_{x \sim \pi_{\theta}(\cdot)} \frac{P(x)}{\pi_{\theta}(x)} \nabla_{\theta} \log \pi_{\theta}(x)
\end{aligned}
\end{equation}
where the importance sampling is used to form a gradient descent algorithm with the distribution match computation as the reward. This step successfully exploits the abundant expressivity in the GAM into an auto-regressive model, where the auto-regressive model could serve as a sampler. The policy gradient method assists the sampling of a generator.

In their seminal work, Khalifa et al. \cite{khalifa2020distributional} introduce a novel methodology that harnesses the power of distributional control in the context of conditional text generation using pre-trained language models (LLMs). 
In contrast to the approach proposed by Parshakova et al. \cite{parshakova2019distributional}, this approach leverages importance sampling while replacing the sampling distribution with an optimal distribution. The variable $q$ is subject to updates exclusively when the condition $D_{KL}(p\Vert \pi_{\theta}) < D_{KL}(p\Vert q)$ holds true.

Korbak et al. \cite{korbak2022on} propose to discover and exploit similarities between DPG (Distributional Policy Gradient) and policy gradient. Although the gradient of DPG (Distributional Policy Gradient) cannot be reduced to a policy gradient, the variance reduction technique of policy gradients can be transferred to DPG. 
Korbak et al. \cite{korbak2021energy} applies KL-adaptive distributional policy gradient (KL-DPG) \cite{khalifa2020distributional} on code generation based on pretraining an auto-regressive model. 
Go et al. \cite{go2023aligning} proposes an $f$-DPG algorithm that allows using of any $f$-divergence as an objective to approximate any target distributions. This approach unifies the formulation of RLHF and DPG (Distributional Policy Gradient). 
The reward of optimization is defined as the negative gradient of $f$-divergence between two distributions $f^{'}(\frac{\pi_{\theta}(x)}{p(x)})$. 

\subsection{Neural Architecture Search}
Previous subsections introduce different purposes of reinforcement learning, bridging the gap of non-differentiable learning systems, incorporating new training signals, and serving as a sampler. Interestingly, the neural network architecture itself can be viewed as a sequence of tokens, therefore being the subjective reinforced generator. Uniquely, the agent itself is a generator but can be applied to almost all feasible tasks which employ neural networks as learners. In this sense, we include Neural Architecture Search (NAS) in this survey even though most applications of NAS are classification tasks.

NAS is used for optimizing neural network architecture \cite{zoph2016neural,pham2018efficient,hsu2018monas,pham2018towards,guo2019irlas,zhong2018practical,tian2020off,chen2020catch,baker2016designing,pang2021rl,chauhan2023dqnas}.
Therefore, the reward for NAS is usually the task metric. For example, when an agent is optimizing the architecture of a classifier, the accuracy of the classifier is usually used as reward \cite{zoph2016neural,hsu2018monas,pham2018towards,guo2019irlas}.
Zoph and Le \cite{zoph2016neural} proposes to use reinforcement learning to guide neural architecture design. They employ an RNN network to generate the architecture description with REINFORCE \cite{williams1992simple} 
ENAS \cite{pham2018efficient} improves the efficiency by parameter sharing.  Instead of building a network from scratch, it constructs the network on pre-defined convolutional cells. This can reduce the search space.
Similarly, MONAS \cite{hsu2018monas} 
It removes states in the training and just considers actions and rewards. It takes power consumption into reward functions. Various optimization goals such as mixing, threshold, and surrogate metric are considered in the reward computation.
IRLAS \cite{guo2019irlas} incorporates inverse reinforcement learning into the NAS.  It defines a feature count that maps an architecture into a trajectory of the agent, $\mu = \sum^{T}_{t=1}\gamma^{t} \phi(s_t)$, where $s_t$ is the architecture information, $\gamma$ is the discount factor, $\phi()$ is the embedding function. They use the $\mu$ to create a linear model for the mirror stimuli function that aims to use the topology of the expert model such as ResNet \cite{he2016deep} as the guidance.

State space and action space are carefully designed in NAS in order to make the training tractable. Layer parameters that are composited as the element of state and action space is a common choice \cite{baker2016designing,rijsdijk2021reinforcement,zhong2018practical}.
MetaQNN \cite{baker2016designing}  constrains the space of states and actions to make the generation tractable. 
Rijsdijk et al. \cite{rijsdijk2021reinforcement} applies MetaQNN on the NAS for side-channel-analysis. The reward function is defined considering the guessing entropy with a different number of attack traces. 
BlockQNN \cite{zhong2018practical} employs neural model generation for image classification tasks. It defines a Network Structure Code (NSC) which quantifies the architecture information such as layer index, operation type, kernel size, and other related nodes in the computational graph. 
 The state of E2GAN \cite{tian2020off} is the average value of each sub-module. The action will be how to extend the architecture,  

The sampling efficiency is also explored. Meta-learning \cite{chen2020catch,pang2021rl} and one-shot learning \cite{chauhan2023dqnas} are two examples.
CATCH \cite{chen2020catch} employs a meta-reinforcement learning framework to accelerate architecture design on meta-testing tasks. 
RL-DARTS \cite{pang2021rl} uses meta-learning as well. 
The meta-optimizer defines the gradient and a control hyperparameter as the state, the shift of the control hyperparameter as the action, and the performance on a valid dataset as the reward to meta-control the direction of the architecture searcher.
DQNAS \cite{chauhan2023dqnas} combines RL-based NAS with one-shot training to get better performance. The key is using one-shot training to transfer weights from some layers that are common to quickly set up the training. 

\section{Application}
Reinforcement learning has been applied to abundant areas. In this section, we organize and classify the literature according to applications, aiming to provide readers a brief introduction of how RL is applied on different areas. The following sections include natural language processing, code generation, computer vision, speech generation, music generation, AI for science and other small areas.

\subsection{Natural Language Processing}
Natural Language Processing (NLP) is one of the largest application areas of generation models and reinforcement learning. 
Reinforcement learning is widely employed in various NLP tasks, like text summarization, machine translation, dialog, etc. We primarily introduce the application directions but may not cover all directions.


\begin{table} 
\centering
\caption{Methods in Neural Language Processing}
\begin{tabular}{cc}
    \toprule
    Sub-areas & Related Works \\
    \hline
    Text Summarization & \cite{paulus2018a,wang2018reinforced,wu2018learning,narayan2018ranking,chen2018fast,gui2019neural,wu2021recursively,yang2020hierarchical,jie2023prompt} \\
    Machine Translation & \cite{wu2018study,pham2018towards,kumari2021sentiment,zhang2017sentence,lam2018reinforcement,zhao2020balancing,luo2019towards,anuchitanukul2022surf,zhao2020reinforced} \\
    Dialog System & \cite{ouyang2022training,serban2017deep,yang2017personalized,sun2023replicating,williams2017hybrid,lewis2017deal,li2017end,ye2022structured,martin2022learning,jang2022gpt,zhong2022hierarchical,saleh2020hierarchical,yang2020improving,srivastava2023response} \\
    Human Value Alignment and Constraints & \cite{liu2022aligning,zhang2020reward,aghakhani2018detecting} \\
    Text, Queries, and Knowledge Graph & \cite{mohankumar2021diversity,dognin2021regen,wang2020grl,wang2021ordering} \\
    Large Language Model & \cite{stiennon2020learning,ouyang2022training,bai2022training,kumar2024training,shani2024multi,zhou2024archer,alabdulkarim2021goal,wu2023fine,jha2024rlsf,chen2024improving,li2024reinforcement,li2024optimizing,ahmadian2024back,li2023remax,richemond2024offline,rafailov2023direct,baheti2023leftover,wulfmeier2024imitating,korbak2023pretraining,snell2022offline,hu2023aligning,deng2022rlprompt,zhang2022tempera,huang2024enhancing,sun2023offline,nottingham2023selective,shen2023loose,dai2023safe,mcintosh2024inadequacy,wang2024rlhfpoison,wang2024reinforcement} \\
    Other NLP Applications & \cite{li2018generative,xu2018diversity,akyurek2023rl4f,wu2022automatic,upadhyay2022efficient,siddique2020unsupervised,wang2024esrl} \\
    \bottomrule
\end{tabular}
\label{tab:application-nlp}
\end{table}

\subsubsection{Text Summarization}
Text summarization  is the process of automatically generating or extracting summaries from a given input document without losing important information. RL has been widely applied in this task by generating the summary \cite{paulus2018a,wang2018reinforced} or extracting the summary \cite{wu2018learning,narayan2018ranking,chen2018fast,gui2019neural}.
Paulus et al. \cite{paulus2018a} proposes to incorporate self-critical policy gradient \cite{rennie2017self} into text summarization. It uses ROUGE as a reward and trains the NLL objective and RL objective by a weighted sum operation.
Wang et al. \cite{wang2018reinforced} change the model to a convolutional sequence-to-sequence model for automatically abstractive topic summarization. 
ROUGE is also utilized in extractive summarization like Wu and Hu \cite{wu2018learning,narayan2018ranking}. 

RL can be used as a means of hard attention mechanism which the policy outputs a discrete action to select contents, unlike in soft attention where a neural network outputs an attention vector that is multiplied with the content. For example, selecting sentences from a document requires hard attention.
Chen et al. \cite{chen2018fast} propose to combine RL as hard attention on the sentence level in the text summarization task. In this work, an agent is a selector who chooses sentences that are valuable to do summarization. The state is the set of documents and the last selected sentence. The action is to choose the next document sentence for extraction. 
VTMRL \cite{gui2019neural} uses reinforcement learning as a hard attention mechanism to filter the less topic-coherent and background words in the task of topic modeling. 
The reward is defined as a sum of a coherence score and a topic overlapping value.

Using RL methods for fine-tuning also attracts attention \cite{wu2021recursively,yang2020hierarchical,jie2023prompt}.
Wu et al. \cite{wu2021recursively} summarize books in a recursive way. This work uses RLHF as training method. The key novelty is a book is summarized based on chapters, then these summaries are fed into the model to do the summarization again to acquire the final summary of a book. Recursive summarization could help human to quickly judge the quality of the summaries.
\cite{yang2020hierarchical} uses policy gradient to adjust a pre-trained abstractive text summarization model. The reward comes from two discriminators. One of the discriminators classify the generated text and human-written text. The other discriminator constructs a ranking loss to motivate the summarizer produces higher probability on human-written summaries.
\cite{jie2023prompt} controls the length of generated texts by PPO-based fine-tuning on text summarization tasks. The reward function is rule-based and calculates the difference between the generated texts and the predefined length limit including target length, upper bound length, and lower bound length.

\subsubsection{Machine Translation}
Reinforcement learning is applied to neural machine translation to bridge the train and test metric gap \cite{wu2018study,pham2018towards}, to direct translation with sentiment preserved\cite{kumari2021sentiment}, to simplify sentences \cite{zhang2017sentence}, to balance human efforts in the interactive system \cite{lam2018reinforcement,zhao2020balancing}, to change the sentiment \cite{luo2019towards}, to diversify the translation output \cite{anuchitanukul2022surf}, or to guide curriculum design \cite{zhao2020reinforced}.
Wu et al. \cite{wu2018study} uses BLEU as a reward in a systematic comparison of decision factors for RL-based NMT.
Pham et al. \cite{pham2018towards} plus BLEU with a HIT reward that counts the coverage of translated results and the annotations into the guidance of training.
Kumari et al. \cite{kumari2021sentiment} applies RL in sentiment preserved review translation. 
The model adopts actor-critic algorithms.
The reward is constructed on a content preservation SBLEU and a sentiment reward that is a dot product of the sentiment vector and output probability distribution. 
DRESS \cite{zhang2017sentence} proposes to employ the policy gradient on the sentence simplification tasks by introducing three desirarta simplicity, relevance, and fluency as rewards.
BIP-NMT \cite{lam2018reinforcement} adopts an actor-critic algorithm for translation model training, which uses a threshold of action entropy for human feedback acquisition and simulate human feedback for evaluation.
Zhao et al. \cite{zhao2020balancing} uses actor-critic. The reward is BLEU and the negative of the times of request for human feedback.  It regularizes the policy with MLE guidance of both the right tokens and human feedback. 
Luo et al. \cite{luo2019towards} use a similar reward formulation with a harmonic mean of two rewards to encourage the model to improve both sentiment reward and content preservation (coherence) reward. 
SURF \cite{anuchitanukul2022surf} defines a new reward function and explores the asynchronous training framework. The reward is defined as 
    $w_F \times e^{F(\hat{y}_{1:t})} + w_S \times e^{SLSS(\hat{y}_{1:t}, X)}$
 Sentence Fluency and Sentence-level Semantic Similarity (SLSS). Sentence Fluency is an average log-likelihood of probabilities given by a pre-trained model. SLSS is a cosine similarity between embeddings of input and generated sentences.
\cite{zhao2020reinforced} uses RL to guide the curriculum design for the machine translation task. The action is selecting a batch of samples from the exposed training dataset. The state is a feature vector that contains the samples for selection and measurements that reflect the performance of the machine translation model like sentence-level log-likelihood. The reward is the performance improvement on the validation dataset.


\subsubsection{Dialog System}

Dialog systems or conversational agents are a complicated but fast-developing area in recent years. The influential InstructGPT \cite{ouyang2022training} is trying to tackle the difficult open-domain dialog system \cite{serban2017deep,yang2017personalized,jaques2020human,sun2023replicating}. Apart from it, a task-oriented dialog system is a parallel but important topic \cite{williams2017hybrid,lewis2017deal,li2017end,ye2022structured,jang2022gpt,zhong2022hierarchical,saleh2020hierarchical,yang2020improving}. Reinforcement learning has been explored in both. Also, a hybrid dialog system emerges recently \cite{srivastava2023response}.

MILABOT \cite{serban2017deep} tackled the Amazon Alexa Prize competition to learn an open-domain chatbot. 
RL is used for model selection. The reward function is a linear regressor trained from collected ratings. 
PRG-DM \cite{yang2017personalized} fine-tunes two policies to generate posts and responses for personalized response generation by policy gradient. 
\cite{jaques2020human} tests the open dialog system by interacting live with human. The system is trained with offline Q-learning with KL control to regularize the policy's action from generating unrealistic language sequences.
Sun et al. \cite{sun2023replicating} propose an imitation learning approach for complex dialogue agents. The imitation objective is defined by Donsker Varadhan's representation of KL divergence to ease the hard problem of high dimensional optimization.

HCN \cite{williams2017hybrid} explores training a dialog control agent with reinforcement learning. To avoid degenerated actions, it iteratively trains with policy gradient and supervised loss. 
Lewis \cite{lewis2017deal} introduce RL on a negotiation task where two agents both have a set of objects and try to exchange objects to make each type of object should belong to one agent. The reward computes whether an agreement is met. 
Yarats \cite{yarats2018hierarchical} apply a hierarchical generation framework and substitute the agent's state from text tokens to latent variables to improve the effectiveness of long-term planning in this game.
Li et al. \cite{li2017end} trains a DQN agent to do task completion in neural dialogue systems. An example is to ask an agent to book a ticket. 
Jaques et al. \cite{jaques2019way} integrates implicit human preferences into a hierarchical open-domain dialogue generation by reinforcement learning. It employs an off-policy batch RL approach with dropout-based uncertainty estimates.
Co-Gen \cite{ye2022structured} proposes to match the latent space of actions in the external database and natural language response in conversational search. 
It uses RL to fine-tune the pre-trained language model with BLEU as a reward. This fine-tuning helps the model achieve better results.
TrufLL \cite{martin2022learning} uses RL for answering questions. Unlike the normal fine-tuning method, it proposes using a pre-trained language model as a truncation module that takes the action space of the agent as input to make the decision-making in a large action space feasible.
\cite{jang2022gpt} applies offline RL (Behaviour Cloning) in dialog agents. The reward is given by an external program which checks whether the user goal is satisfied. This work trains on a dataset that includes the self-generated action candidates which are evaluated by a critic network to address the sparse and high-dimensional action space.
\cite{zhong2022hierarchical} adopts a hierarchical RL for the medical dialog system. The reward motivates the agent to collect more information about the disease and produce a successful diagnosis.
\cite{saleh2020hierarchical} proposes a hierarchical framework in which a high-level policy determines the semantics at the utterance level that is in the latent space and a low-level policy that outputs the next word. The reward is designed to minimize repetitiveness and toxicity and enhance consistency and sentiment.
\cite{yang2020improving} incorporates A3C \cite{mnih2016asynchronous} into negotiation dialog systems. The reward motivates the success of reaching a compromise and penalizes the result that no deal is struck, incentivizing both agents to negotiate and achieve a both preferable result.

READER \cite{srivastava2023response} is designed for mental health counseling agents by generating dialogue in a hybrid way. The agent needs to understand various contents from users but provides effective and appropriate responses. 

\subsubsection{Human Value Alignment and Constraints}
The output of generators is not well matched with human values. Models sometimes have hallucinations, generating fake information that they do not understand at all. Sometimes models are impacted by datasets and spit out sentences that do not match human values in some cultures. Reinforcement learning can be used to adjust the model to work better on value matching \cite{liu2022aligning}, impose constraints \cite{zhang2020reward}, or even help people to combat problems like fake news \cite{aghakhani2018detecting}.
SENSEI \cite{liu2022aligning} proposes to use actor-critic to align text generation with human judgments. The reward is predicted by a binary classifier trained a human-labeled text data. 
RCR \cite{zhang2020reward} 
uses a discriminator to model the violations and computes a penalty accordingly. This penalty is added to the reward to regulate the actions of the text generator.
FakeGAN \cite{aghakhani2018detecting} trains a deceptive reviews classifier with a two-discriminator GAN model. Although it addresses a classification problem, the method contains generating deceptive reviews, which is a generation subtask. 

\subsubsection{Text, Queries and Knowledge Graph}
Natural language is a good interface for humans but not for search engines and databases. Therefore, translating natural language into structured queries and knowledge graphs poses a long-term problem in NLP. Reinforcement learning is used in these applications to optimize the query quality \cite{mohankumar2021diversity}, create knowledge graphs \cite{dognin2021regen}, complete them \cite{wang2020grl}, and even causal graphs\cite{wang2021ordering}.
Mohankumar et al. \cite{mohankumar2021diversity} incorporate human preference rewarded RL in query rewriting for advertising. The reward is provided by a fine-tuned model based on another model pre-trained in 100 languages. 
ReGen \cite{dognin2021regen} proposes to use reinforcement learning to guide the text and knowledge generation based on large pre-trained models. 
GRL \cite{wang2020grl} integrates GAN and RL in the knowledge graph completion. The model is constructed on graph neural networks and LSTM. 
The state space includes a combination of both entity space and relation space in the graph. The action is to select the neighboring relational path to extend the path.
CORL \cite{wang2021ordering} proposes to generate a causal graph by reinforcement learning. The state is defined as an embedding of a causal graph node, the action is also in the node space but takes order constraints into account by imposing a mask to force the parent nodes chosen from a certain set. The reward is defined as Bayesian Information Criterion and the episodic reward and dense reward settings are explored.

\subsubsection{Large Language Models}
Recent years have witnessed a big rise of large language model \cite{stiennon2020learning,ouyang2022training,bai2022training}. We collect and organize advances of RL applications in this direction.
Recently, to further enhance the capability of the large language model, multi-turn reinforcement learning is proposed \cite{kumar2024training,shani2024multi,zhou2024archer}, where the reward is not spontaneous in contrast to single-turn RL.
\cite{kumar2024training} incorporates self-correction into LLM. To make the agent revise its action, the action (generated sequence) is piped into itself to learn a correction step. The reward is derived from human preferences over two multi-turn conversations. It adopts a two-stage tuning scheme to model the two attempts of the inference. The first attempt produces a response, then revises or corrects the response in the second attempt. The first stage pushes the second attempt for correction and keeps the first attempt frozen. The second stage jointly optimizes the two attempts with reward shaping to penalize falsely changing the first attempt.
\cite{shani2024multi} addresses the limitation of single-turn RLHF by introducing a multi-turn algorithm that takes the history of the multi-turn interactions as states, outputs the response as actions, and the reward is generated by human preference. GAE \cite{schulman2015high} is used to optimize the policy and KL divergence is leveraged to constrain the optimization near the original model to prevent policy collapse.
\cite{zhou2024archer} transfers hierarchical RL into mutli-turn langugage model training. The state space and action space are similar to \cite{shani2024multi}. The reward is the success of long-term objective. For example, the agent is asked to search a book online, the success signal would be 1 if the book is searched. The high-level action is utterances in response of the state, and the low-level is fine-grained response of the high level action. The model could be combined with either online RL algorithms or offline RL algorithms.

New reward formulations are proposed to enhance the effectiveness \cite{alabdulkarim2021goal,wu2023fine,jha2024rlsf,chen2024improving,li2024reinforcement,li2024optimizing}.
\cite{alabdulkarim2021goal} uses a language model and a reinforcement learning agent to generate stories towards specific goals. The language model is responsible for generating multiple goal-conditioned continuations of a story which are selected by the knowledge graph-based reinforcement learning agent. The selected sequence will be appended after previous contexts for next continuation generation. The reward caters for goal achievement and coherence reward.
\cite{wu2023fine} investigates the effectiveness of fine-grained reward for RLHF. The responses generated by an LLM is segmented into sub-sentences. Fine-grained reward entails human labeling these sub-sentences on the exact problems of the LLM response, such as inveriable facts, irrelevance, or information incompleteness.
\cite{jha2024rlsf} proposes RLSF which exploits the symbolic feedback system to guide the fine-tuning of an LLM. A certificate of each response the LLM produces is generated by a symbolic reasoner, which is fed into the reward model for PPO-based fine-tuning. This work applies the method to the task that transforms a psuedo-code to C++ code.
\cite{chen2024improving} proposes a token-level reward model unlike common instance-level reward model for the LLM tuning. The token-level capability is acquired by distilling a generative correction model that generates a probability in deciding a reward for each token. The reward model is combined with PPO to fine-tune the LLM.
\cite{li2024reinforcement} uses token-level feedbacks for LLM fine-tuning. The reward function provides token-level reward by computing the difference between probabilities before and after the word is generated. To increase robustness, this work also uses a "First quantize then noise" strategy that uses quantized rewards and inject noise into rewards but keep them in the interval.
\cite{li2024optimizing} composes an algorithm that can balance multi-objective during PPO-based fine-tuning. The reward function is a weighted average of all specific rewards. The method uses mirror descent and smooth to update the weights which guarantees that mitigates over-focus on single objective. This method is used to make the reward composition fair and stable.

Apart from reward function, new advances in objective functions \cite{ahmadian2024back,li2023remax,richemond2024offline,rafailov2023direct,baheti2023leftover,wulfmeier2024imitating,korbak2023pretraining,snell2022offline,hu2023aligning}, reflection capability \cite{shinn2024reflexion,pang2023language}, scaling \cite{havrilla2023trlx}, ensemble \cite{zhang2024improving} are proposed to further improve large language models.
Similar to RLOO \cite{ahmadian2024back}, ReMax \cite{li2023remax} is proposed to leverage a variant of REINFORCE to substitute PPO \cite{schulman2017proximal} in RLHF \cite{stiennon2020learning}.
\cite{havrilla2024teaching} proposes to enhance the reasoning capability of LLM on techniques like Expert iteration, Return-conditioned RL, and outcome-based reward modeling.
DRO \cite{richemond2024offline} is proposed to exploit single-trajectory dataset in LLM fine-tuning. Single trajectory dataset means that one data point consists of a prompt, a response, and human preference. In contrast, DPO \cite{rafailov2023direct} requires each data point to have two responses for preference learning. DRO achieves this by changing the objective by a MSE loss between reward, value function, and the KL divergence regularization term.
\cite{baheti2023leftover} fine-tunes an LLM with an advantage-based offline RL on pre-existing dataset. The LLM is trained with the importance sampling on data collected by a reference LLM. To increase the robustness of the fine-tuning, this work only selects data point with a positive reward for training.
\cite{wulfmeier2024imitating} fine-tunes LLMs with Inverse RL. This work creates a link from inverse soft Q-learning to a regularized MLE objective, which enables to trade-off the impact of regularization that focuses more on long-term impact of reward on action sequence, increasing the diversity of the action.
\cite{korbak2023pretraining} explores human preference learning in the pre-training stage instead of common fine-tuning stage. It tests five objectives of pretraining, including Reward Weighted Regression (RWR) and Advantage Weighted Regression (AWR). RWR a variant of policy gradient method. AWR substitutes the reward with estimated advantages.
\cite{snell2022offline} proposes to incorporate offline Q learning into the language generation learning. This work combines the utility maximization of reinforcement learning and stability of the supervised learning by Bellman backups of a value function and a Q function, forming an implicit value function learning. The token is generated based on the history of all tokens instead of individual transition. The reward is based on downstream tasks.
\cite{hu2023aligning} explores fine-tuning the LLM by three different offline reinforcement learning algorithms which can save computing resources compared to PPO. The first method selects high quality response and drop others. The second method is similar to policy gradient with exponentiated rewards. The third method adds reward in the prompts for better alignment.
Reflexion \cite{shinn2024reflexion} proposes verbal reinforcement learning which does not train or fine-tune the LLM but using the in-context learning ability to form a reinforced loop to achieve the correct answer for a prompt. The key insight is creating a reflection loop, where an actor language model gets observations (prompts) from the environment and reflections from a memory buffer which adds verbal reflections of actions and outcomes iteratively. The outcome is generated by an evaluator language model and is transformed to reflections in verbal form by a self-reflection language model. The actor language model try multiple times until the answer is correct or the maximum trial limit is reached.
Similarly, \cite{pang2023language} leverages the contemplation ability of a LLM. This work shows that the LLM is good at self-evaluation of its own generated texts. Therefore, the self-evaluation is used to assign rewards for further reinforcement learning fine-tuning.
\cite{havrilla2023trlx} addresses problems of scaling RLHF to large models up to 70B (70 billion parameters). This work employs distributed training strategies entails data parallelism, model sharding, and pipeline parallelism to efficiently train the large-scale models. The result shows that PPO is effective but sensitive to hyper-parameter settings. Offline RL method is easier to train but achieves sub-optimal performance.
\cite{zhang2024improving} proposes to integrate model ensemble into the reward model of RLHF. This work discusses three types of reward model ensembles: single reward model ensemble, linear-layer ensemble, and Low-Rank Adaptation (LoRA)-based ensemble. The single reward model ensemble uses multiple reward models directly. In the linear-layer ensemble, reward models share the same Transformer model and predicts different rewards on a single linear layer. LoRA-based ensemble integrates a LoRA layer before the single linear layer. This investigation aims to efficiently compute the reward model for scaling of large language model fine-tuning.

RL are also used in prompt optimization \cite{deng2022rlprompt,zhang2022tempera,huang2024enhancing,sun2023offline,nottingham2023selective}.
\cite{deng2022rlprompt} employs RL for prompt optimization. The state space is an initial prompt and corresponding task description. The action space is to select or modify the prompt by generating prompts. The reward is defined by the downstream tasks and is normalized to stablize the training.
\cite{zhang2022tempera} utilizes RL for prompt editing in test-time. This method uses the last hidden states of the pretrained language model as state representations, the LLM can choose the objects from instruction, in-context examplars, and verbalizers. The reward formulation uses the same one as in \cite{deng2022rlprompt} but considers difference between edits to motivates the LLM to accumulates reward at each edition.
Similarly, \cite{huang2024enhancing} refines prompt towards truthful, benign, and helpful outputs of target LLM by reward formulation. This work employs open-source models and publicly available dataset to construct reward models for quality, safety, and jailbreak prompts.
Prompt-OIRL \cite{sun2023offline} proposes a query-dependent prompt optimization approach, which is different from \cite{deng2022rlprompt,zhang2022tempera} that both search for distributional optimal prompt (expected quality of answers). \cite{sun2023offline} seeks a single best prompt. Besides, this work combines an offline reward model for inference time evaluation to save the cost of interacting with LLM for rewards.
In \cite{nottingham2023selective}, reinforcement learning is utilized to optimize the input space of an LLM. This method separates the input of an LLM into task description and individual features. An reinforcement learner is created for selecting the optimal set of individual features to guide the LLM towards right responses. The output of the reinforcement learner is integrated  into the LLM for downstream tasks. The reward is formulated as the action likelihood of the LLM when the individual feature is in the valid subset.

Safety concerns are discussed in RL fine-tuned LLM \cite{shen2023loose,dai2023safe,mcintosh2024inadequacy,wang2024rlhfpoison,wang2024reinforcement}.
\cite{shen2023loose} demonstrates that the long response bias could be mitigated by Product-of-Experts, a reward function is proportional to a product of the human preference reward and the length bias reward. A bias-only expert is introduced to capture the length bias reward. The  bias-only expert is trained with perturbations to decrease the impact of semantics.
\cite{dai2023safe} proposes SafeRLHF to balance helpfulness and harmlessness in LLM alignment. The reward model is separating helpfulness and harmlessness by taking harmlessness as a constraint embedded in the Lagrangian optimization. The reward models are trained separately with different datasets based on preference-based function and are fused during PPO-based fine-tuning. The Lagrangian multiplier is updated by a moving average of harmlessness cost function.
\cite{mcintosh2024inadequacy} demonstrates that RL fine-tuned LLM still suffers from semantic vulnerabilities where the radicalized response could be elicited with designed prompts.
\cite{wang2024rlhfpoison} shows that LLM might be attacked without compromising original safety alignment objective. This work attacks an LLM by data poisoning. Specifically, it flips the label of the data for RLHF's reward function but filters out poison data that could induce significant changes. The attacked model generates longer response when specified trigger word appears.
\cite{wang2024reinforcement} attacks the LLM with an RL trained LLM, where the attacker LLM is trained to search for jail-break prompts that leads the innocent LLM towards targetted attacked behaviour. This attack process could be used for Trojan detection by attacking the target LLM and expose the sensitive prompts.

\subsubsection{Other Applications in NLP}
Reinforcement learning application in NLP is a wide area but we focus on typically generating sequences for other purposes like review generation\cite{li2018generative,xu2018diversity}, critique generation \cite{akyurek2023rl4f}, mathematical problems generation \cite{wu2022automatic}, keyword-to-sentence generation \cite{upadhyay2022efficient}, paraphrasing \cite{siddique2020unsupervised}, or mutiple tasks \cite{wang2024esrl}.
Li et al. \cite{li2018generative} combines adversarial training and reinforcement learning for review generation of commercial purposes. It uses RL in the same way as SeqGAN\cite{yu2017seqgan} do.
DP-GAN \cite{xu2018diversity} applies similar framework like in \cite{li2016deep,yu2017seqgan}. It separates the reward into two levels, word level, and sentence level. The sentence level reward is an average of all rewards of the word in it. The total reward is a product of sentence-level reward and the discounted sum of word-level reward.
RL4F \cite{akyurek2023rl4f} proposes to enhance the large language model by critique generation. 
An critique LLM is used to generate critique for the task LLM. 
MWPGen \cite{wu2022automatic} generates mathematical problems from a math expression and some topic words. It does the problem generation and then uses the generated problems to get an answer by a neural network-based solver. It defines the reward to check the correctness between the expression generated by the solver and the original expression. 
Upadhyay et al. \cite{upadhyay2022efficient} redirect a pre-trained language model towards multiple rewards to improve performance. 
It studies unsupervised controlled text generation and takes text style into consideration. 
\cite{siddique2020unsupervised} applies RL fine-tuning in unsupervised paraphrasing tasks. The method entails three steps, pre-training, transition, RL tuning. In pretraining, a VAE objective is used. In the transition phase, the agent learns to generate longer sequence without supervision. In the RL phase, the agent is trained to maximize the reward that optimizes the output towards semantic adequacy, language fluency, and expression diversity.
\cite{wang2024esrl} investigates improving the memory and time efficiency for RL-based sequence generation. The reward is task-specific, including BLEU \cite{papineni2002bleu} and ROUGE \cite{lin2004rouge}. It improves the efficiency by reducing redundant sampling and decreasing the size of computational graphs. This work conduct experiments on machine translation, text summarization, and RLHF.

\subsection{Code Generation}
Given the application of RL on NLP, it is also natural to consider if the coding process can be automatically executed by machines. Within them, RL-based models perform well and improve their performance in multiple directions.

\begin{table}
\centering
\caption{Methods in Code Generation}
\begin{tabular}{cc}
    \toprule
    Sub-areas & Related Works \\
    \hline
    Code Search & \cite{yao2019coacor,wang2022enriching,dai2024enhancing} \\
    Comment and Annotation Generation & \cite{wang2020reinforcement,cai2020tag} \\
    Code Generation & \cite{wang2022compilable,zhang2022learnedsqlgen,le2022coderl,shojaee2023execution,duan2023leveraging,liu2023rltf,gehring2024rlef,dou2024stepcoder} \\
    Unit Test Generation & \cite{steenhoek2023reinforcement} \\
    \bottomrule
\end{tabular}
\label{tab:application-code}
\end{table}

\subsubsection{Code Search}
Code search takes natural language text as input, and searches for a code snippet that can solve the problems presented by the text. RL is implemented to use annotation \cite{yao2019coacor} or enhanced query \cite{wang2022enriching} for code search.
CoaCor \cite{yao2019coacor} applies RL in code annotation, a form of code generation for code retrieval. They use an actor-critic algorithm where states are code snippets, actions are generated code, and rewards are defined according to code retrieval requirements.
QueCos \cite{wang2022enriching} uses the policy gradient method in the code search application. The agent learns to generate queries to get high-quality matched code snippets. The agent is guided by a ranking reward and a BLEU reward. 
\cite{dai2024enhancing} applies RL-based fine-tuning in query rewriting for 
E-commerce applications. The LLM takes a user query and pattern as input, and rewrites the query to better align with the product catalog's terminology. The capability of the LLM enhances the semantics understanding of the query. The LLM does Supervised Fine-tuning (SFT) and then uses RL to fine-tune for better search performance measured by a relevance reward that measures the relevance between the query and the rewrited version, a productive reward measures effectiveness and an increment reward measures what the rewriting changes the result. 

\subsubsection{Comment and Annotation Generation}
Reinforcement learning can be used for code summarization as well \cite{wang2020reinforcement,cai2020tag}, which takes code as input and outputs natural language to summarize the usage of the code.
Wang et al. \cite{wang2020reinforcement} proposes to use RL to guide code summarization with a hierarchical attention network. It uses RL to combat the exposure bias \cite{wan2018improving} in the code summarization dataset. The reward is BLEU metric, and the algorithm is actor-critic architecture.
TAG \cite{cai2020tag} incorporates type auxiliary guiding for code comment generation by reinforcement learning. It contains two stages in the decoding process, an operation selection stage, and a word selection stage. RL is used to guide the operation selection stage because there is no labelled signal to learn it directly. Similarly, it uses non-differentiable evaluation metrics to provide rewards and trains the two stages jointly under the RL framework.

\subsubsection{Code Generation}
Code generation is different from code search in the sense that it directly generates the code instead of searching based on matching.
COMPCODER \cite{wang2022compilable}'s
training consists of three stages. In the first stage, the code generation model is fine-tuned from a language model. Then, reinforcement learning is used to introduce the compiler guidance. The last stage uses a discriminator to learn the compiler feedback on the generated candidates. The discriminator is trained on whether the code can be successfully compiled. 
LearnedSQLGen \cite{zhang2022learnedsqlgen} applies actor-critic algorithm on SQL generation problem. The state contains elements of SQL sequence, including reserved words like \textit{Select, From, Where}, metadata of tables and attributes, cell values, operations like $=,>,<$, and EOF showing the termination of the sequence. The action space is the same as the state space. 
CodeRL \cite{le2022coderl} incorporates unit tests into code generation. The code generator is trained with rewards that are defined based on unit test signals that are ${CompileError, RuntimeError, FailedTest, PassedTest}$. A critic network is used to predict the probability of four types of test signals.
PPOCoder \cite{shojaee2023execution} generates code with multiple constraints in reward. The reward function is defined as a sum of test error, syntactic matching score, semantic matching score, and a KL constraint to prevent RL from diverging too far. 
\cite{duan2023leveraging} combines a preference-based learning framework into code optimization. The method integrates unit test feedback for reward function to improve correctness and efficiency of LLM fine-tuning.
\cite{liu2023rltf} proposes to incorporate unit test as feedback as well. The reward function comprises three terms, coarse reward, fine reward, and adaptive reward. The coarse reward assigns scalar reward according to the result of the code: pass, failure, syntax error, and other errors. The fine reward assigns a number according to the specific error types. The adaptive reward takes the number of passed test cases into account and stimulates the LLM to pass as much test cases as possible.
\cite{gehring2024rlef} uses execution result as reward as well, the difference is that the reward function returns high reward only when all private test cases are passed.
\cite{dou2024stepcoder} proposes a curriculum learning for the code generation. The agent is required to generate a small amount of code snippet given other parts canonical solution. The curriculum gradually increases difficulty by decreasing the portion of accessible canonical solution and increasing the length of generation.

\subsubsection{Unit Test Generation}
\cite{steenhoek2023reinforcement} improves the quality of unit test generation by PPO-based fine-tuning. The state consists of the current code under test and the unit test generated by the LLM. The later allows the LLM to inspect and refine its generated results. The action is generating the unit test. The reward is acquired from a static quality analyzer and is used to train a reward model that provides rewards for PPO-based fine-tuning.



\subsection{Computer Vision}
Computer vision is another cornerstone of modern machine-learning research. Generation tasks in computer vision are also capable of using reinforcement learning algorithms in many sub-areas, including text generation tasks such as image captioning, visual question answering, visual dialog, and visual entity generation like image generation and 3D objects and scenes generation.

\begin{table} 
\centering
\caption{Methods in Computer Vision}
\begin{tabular}{cc}
    \toprule
    Sub-areas & Related Works \\
    \hline
    Image Captioning & \cite{rennie2017self,ren2017deep,zhang2017actor,miao2020multi,shen2020remote,shi2021partial,nie2021triangle,dessi2023cross} \\
    Visual Question Answering & \cite{zhang2018goal,zhao2020open,saqur2020multimodal,liu2020cascade} \\
    Visual Dialog System & \cite{fan2020recurrent,zhang2018sch} \\
    Text-to-Image Generation & \cite{black2023training,fan2023dpok,bai2022constitutional} \\
    3D Generation & \cite{akizuki2020generative,ostonov2022rlss,zhang2022qinet,zhang2023point,lin2020modeling,zhao2023synthesizing,siyao2022bailando} \\
    Other Computer Vision Tasks & \cite{khan2024self,deng2022dashbot,liu2022video,yu2023fusing,zhai2024fine} \\
    \bottomrule
\end{tabular}
\label{tab:application-cv}
\end{table}

\subsubsection{Image Captioning}
Image captioning is a task where the model aims to describe related events and entities in an image.  New algorithms are explored, including reward normalization \cite{rennie2017self,shen2020remote}, architecture advancement \cite{ren2017deep}, new reward function \cite{zhang2017actor,miao2020multi,seo2020reinforcing,shi2021partial,nie2021triangle,dessi2023cross}.
SCST \cite{rennie2017self} incorporates REINFORCE with a baseline as the training algorithm to normalize the rewards an agent experiences. It defines the reward by the performance of a current model under the inference algorithm. The baseline uses the test dataset to compute the reward.
Ren et al. \cite{ren2017deep} uses an RL algorithm whose state comprises an image and up-to-now generated words. An action is the next word. The reward is defined as a cosine similarity between the generated captions and the image. 
Zhang et al. \cite{zhang2017actor} uses an actor-critic algorithm and a separation of RNN between the actor-network and the critic network to do image captioning. 
TOPIC \cite{miao2020multi} uses policy gradient on multi-model product title compression where text and images are input for title generation. 
\cite{shen2020remote} applies RL fine-tuning in remote sensing image captioning. The RL algorithm uses \cite{rennie2017self}.
OffPG \cite{seo2020reinforcing} implements the reward function with human feedback. The ratings of captions are provided and incorporated into a policy gradient with baseline to learn a captioning system in a offline way.
Shi et al. \cite{shi2021partial} improves the diversity of generated captions by increasing the exposure of varied caption candidate and a reward function max-CIDEr that relaxes the similarity constraint to improve diversity.
\cite{nie2021triangle} proposes a triangle-reward for policy gradient training. Three rewards measure visual consistency, fluency and correctness, and semantic coherence respectively. The first and the last reward is constructed based on graph representations of the scene.
DiscriTune \cite{dessi2023cross} provides reward signal through a pre-trained fronzen image retriever to a REINFORCE-based image captioner. The reward is defined as whether the image retriever gets the original image according to the caption generated by the captioner.

\subsubsection{Visual Question Answering}
In VQA, topics like reward design \cite{zhang2018goal,saqur2020multimodal,liu2020cascade} and new architecture\cite{zhao2020open} are interesting topics.
VQG \cite{zhang2018goal} applies RL on the vision-and-language problems like the environment GuessWhat \cite{de2017guesswhat}. This game comprises three components, a questioner, a guesser and an oracle. The guesser guesses the targeted object in an image given the context collected by the questioner. The oracle contains information about the targeted object. VQG models the questioner as the agent and proposes to formulate the reward through three dimensions: goal achieved reward, progressive reward, and informativeness reward.
Zhao et al. \cite{zhao2020open} combines multi-model representation learning with a reinforced GAN-based model for visual question answering. The representation model contains a pre-trained convolutional network, a frame-level dynamics network, and a segment-level attention network.
MGN \cite{saqur2020multimodal} uses REINFORCE algorithm to fine-tune a graph neural network that takes image and text sequence as input. The state is the image and query, the action is a distribution of tokens that impact a symbolic program which generates the final answer, and the reward is the final answer correctness.
\cite{liu2020cascade} incorporates the policy gradient method into a cascade reasoning-based image captioning framework, allowing the agent to learn through trial and error. The reward is average normalized Levenshtein similar-
ity (ANLS) that measures the similarity between generated answers and the ground truths.

\subsubsection{Visual Dialog System}
RL enhances the visual dialog system by incorporating discriminators \cite{fan2020recurrent,zhang2018sch}.
Fan et al. \cite{fan2020recurrent} borrows SeqGAN \cite{yu2017seqgan} model into a visual dialog system. They devise a model that contains two modules, an encoder to embed images, captions, and questions into the embedding vector, which is fed into an RL-based decoder as the state. The decoder is an RL-based GAN. The generator is the agent that outputs answers as actions. The discriminator learns to classify the generated answers from real ones in the embedding space. 
SCH-GAN \cite{zhang2018sch} learn a cross-modal hashing GAN with reinforcement learning. Text and image modalities are considered. The generator tries to retrieve an image from texts or vice versa. The discriminator aims to distinguish true examples of the query. 

\subsubsection{Text-to-Image Generation}
Recent advancement in image generation is diffusion models, thereby it might be beneficial explore how to combine reinforcement learning with diffusion models by improvement on images characteristics that are hard to be described by prompts\cite{black2023training} and online reinforcement learning methods \cite{fan2023dpok}.
The action for the agent in \cite{black2023training} is the noise vector generated in steps given the last step output and a context variable. 
The agent is first pre-trained on DDPM loss reweighted by exponentiated rewards. The policy gradient algorithms with importance sampling are incorporated to train the agent. The reward takes file size and human aesthetic preference acquired from another predictor into consideration. An extra alignment using a vision language model is incorporated for RLAIF \cite{bai2022constitutional}.
\cite{fan2023dpok} proposes to compare the RL-directed fine-tuning and supervised fine-tuning in the context of KL divergence as a regularizer. 
The RL fine-tuning uses a policy gradient with a KL term to constrain models on a pre-trained model. The reward in RL fine-tuning is typically from human preference matching.

\subsubsection{3D Generation}
Apart from 2D images, 3D generation is able to introduce reinforcement learning to get a new generation method\cite{akizuki2020generative}, better scene generation\cite{ostonov2022rlss}, surface completion\cite{zhang2022qinet}, point clouds completion\cite{zhang2023point}, mesh edition \cite{lin2020modeling}, human motion generation \cite{zhao2023synthesizing,siyao2022bailando}.
Akizuki et al. \cite{akizuki2020generative} propose to use RL to generate objects in 2D or 3D space. It models the generation as a link game, where an agent is required to link from pixel to pixel or from voxel to voxel. The action space is the direction to extend the pixel. The reward is based on whether the next pixel or voxel is in the object. It also successfully learns to generate Lego structures given fabrication constraints.
RLSS \cite{ostonov2022rlss} generates 3D indoor scenes with reinforcement learning. The state contains structures or objects represented by the center position and bounding boxes as well as the replacement information. The action is to place an object in a place. The reward is constructed by programs that include multiple conditions such as successful condition, count of objects, and failure conditions. 
QINet \cite{zhang2022qinet} completes the corrupted 3D point cloud with the actor-critic algorithm. It first generates masks as pre-processing. Then it converts the discrete point cloud to the continuous surface. 
The policy is trained by rewards defined as IoU between generated cloud and the true cloud and a latent code constraint to prevent the latent code drift far away.
Zhang et al. \cite{zhang2023point} complete point cloud with A3C algorithm.  The state is the updated point cloud of each iteration. The action is the next best view for the completion. The agent adjusts the camera points in the coordinate system to change the views. 
\cite{lin2020modeling} proposes to edit 3D shapes with two Double DQN agents, a prim-agent that modifies primitives (e.g., cuboids) and a mesh-agent that changes the vertices on the mesh. The state is the current configuration of the 3D shape. The action is editing the corner of the primitives or deleting primitives for the prim-agent, and is changing groups of vertices. The reward includes the difference between intersection over union (IoU) of primitives before and after taking an action.
\cite{zhao2023synthesizing} employs PPO to synthesize human motions in 3D indoor scenes. The state contains configurations of 3D scene geometry, virtual human body, and intended goals to interact with. The action is defined in a latent space of pre-trained motion generation models. The reward enhances goal-reaching. foot-floor contact, and penetration avoidance.
\cite{siyao2022bailando} generates 3D Dance via a transformer-enhanced actor-critic method. The state encompasses a sequence of dance poses. The action is the next dance pose. The reward motivates the agent to align motion-music beats and penalizes inconsistency movement between upper body and lower body.

\subsubsection{Other Computer Vision Tasks}
Other tasks like visual program synthesis \cite{khan2024self}, dashboard generation \cite{deng2022dashbot}, video summarization \cite{liu2022video}, and vision-language models \cite{yu2023fusing,zhai2024fine}.
Visual program synthesis is a task that takes a query as input, the agent learns to write program for visual tasks by utilizing off-the-shelf computer vision models like image detection. \cite{khan2024self} proposes to address this problem by optimizing the agent with rewards that are binary variables computed based on the existing vision-language annotations (not the target code) to guide the tuning of the language model that generates code.
\cite{deng2022dashbot} proposes to employ RL in dashboard generation that is frequently employed in business intelligence to help data analysts with data exploration. The state represents the current configuration of the dashboard, including selected data fields, chart types, and arrangements. The action is to add a chart, remove one, or change properties of a chart. The reward contains three terms: diversity that captures the number of chart types, parsimony that reveals the number of charts, and insight that shows the number of correlations revealed in the chart.
Liu et al. \cite{liu2022video} propose to use policy gradient in the video summarization task. The state is the spatio-temporal features learned from a representation learning network. The action is selecting frames by binary classification. The reward is increasing similarity of a selected representative frame, aka medoid, and other frames in the same cluster and decreasing similarity between medoids.

Reinforcement Learning could be used in tuning Vision-Language Model (VLM) \cite{yu2023fusing,zhai2024fine} as well. \cite{yu2023fusing} proposes to adapt pre-trained language model for multi-modal tasks by a RL-based training. This method trains a multi-modal encoder and fix other parts of the neural network. The encoder takes outputs of an image from a pre-trained CLIP model and generates multi-modal representations for text generation. The reward is formulated based on consine similarity between CLIP embeddings of the image and the generated text, which measures the alignment between the input image and the generated text. The  training objective also poses a KL penalty to keep the RL policy (text generator) close to the original policy. \cite{zhai2024fine} formulates the VLM as a policy, aiming to improve its effectiveness the decision making capability via PPO-based fine-tuning. The state space consists of image pixels and task descriptions in the text form. The action space is in text form as well. The reward could be derived from the success of the task. For example, in the number line task, the agent is given two poker cards and is required to match the number of the two cards. The reward could be 1 for success, -1 for incorrect actions, and 0 otherwise.


\subsection{Speech and Music Generation}
Speech and music data can be transformed to sequential data points. Thereby, it is natural to incorporate reinforcement learning. In practice,RL is employed to improve the quality \cite{mohan2020incremental,liu2021reinforcement}, decrease latency \cite{mohan2020incremental}, control the bit rate in speech coding \cite{gibson2022reinforcement}, and melody generation \cite{li2021symbolic}.
Tacotron 2 \cite{mohan2020incremental} learns an agent to control the text2speech translation. The state of the agent is a product of an attentive vector and hidden vectors, the attentive vectors, and the output sequence. The action is whether to read or speak. When it reads, it generates an attention vector. When it speaks, it generates the output in the form of a mel-spectrogram frame. The reward motivates the agent to produce high-quality translation as well as low latency.
i-ETTS \cite{liu2021reinforcement} explores reinforced emotional text-to-speech synthesis. The input of the model is reference audio and the targeted character sequence. The reference latent vector encodes tokens that indicate the emotion with an attention model. Then two modality is fused to decode and generate an emotional speech that is fed into a speech emotion recognition classifier. The reward for the speech generator is recognition accuracy. The agent is trained with a policy gradient.
Gibson and Oh \cite{gibson2022reinforcement} apply RL in speech coding, an essential technology for digital cellular communications. 
The agent is a tree-structured controller for the bit rate in speech coding. The system uses a reconstruction error as the penalty. 
\cite{li2021symbolic} proposes to incorporate LeakGAN \cite{guo2018long} into music melody generation. The music notes are converted to symbolic representation. 

\subsection{AI For Science}
Machine learning community also want to help scientists in other research areas with useful machine-learning tools. Molecule design is a critical area because the design process is typically an expensive and long process. Therefore automatically finding patterns from large amounts of data accumulated in scientific areas become another hot area in recent years. RL can also play an important role in its flexibility.

\begin{table} 
\centering
\caption{Methods in AI For Science}
\begin{tabular}{cc}
    \toprule
    Sub-areas & Related Works \\
    \hline
    Molecule Design & \cite{putin2018reinforced,popova2018deep,putin2018adversarial,de2018molgan,goel2021molegular,blaschke2020memory,thomas2022augmented,brown2019guacamol,chen2021fragment,ishitani2022molecular,fu2022reinforced,hu2023novo,sun2022molsearch,liu2023drugex,simm2020reinforcement,zholus2024bindgpt} \\
    Reaction Optimization & \cite{zhou2017optimizing,volk2023alphaflow,gottipati2020learning,rajak2021autonomous,powell2020real} \\
    Micro-structure Generation & \cite{nguyen2022synthesizing} \\
    Quantum Architecture Design & \cite{kuo2021quantum,lin2020quantum,ostaszewski2021reinforcement,moro2021quantum} \\
    \bottomrule
\end{tabular}
\label{tab:application-ai-for-science}
\end{table}

\subsubsection{Molecule Design}
Drug discovery or de novo molecule design can be viewed as policy search in the molecule space. Thereby it is natural to introduce RL on this application. Abundant research has been conducted, including various mainstream methods listed in Section \ref{sec:benefits}, like GAN-based models \cite{sanchez2017optimizing}, human prior \cite{olivecrona2017molecular}.

For Gan-based models, the direction of architecture design \cite{sanchez2017optimizing,putin2018reinforced}, sample selection\cite{putin2018adversarial} are explored.
ORGANIC \cite{sanchez2017optimizing} uses RL directly in the GAN model to discover drugs. 
RANC \cite{putin2018reinforced} combines ORGANIC style architecture with a memory network DNC, which enables the model to remember complex sequences and generate longer sequences.
ReLeaSE proposed by Popova et al. \cite{popova2018deep} 
incorporates a prediction model to bias the generated chemical structures toward those with the desired physical or biological properties.
ATNC \cite{putin2018adversarial} modifies the OGRANITC model by introducing a new sample selection scheme function that filters out molecules which are far from training samples and regenerates the sequences until the number of new sequences is higher than the threshold. 
MolGAN \cite{de2018molgan} uses DDPG on small molecular graph generation to cope with high dimensional action space. The reward is emitted from a differential approximation of the true reward function.

For human-designed reward methods, various topics are investigated, like reward design \cite{olivecrona2017molecular,goel2021molegular,simm2020reinforcement}, memory architecture \cite{blaschke2020memory}, ranking methods \cite{thomas2022augmented}, training methods \cite{brown2019guacamol}.
Inspired by Sequence Tutor \cite{jaques2017sequence}, REINVENT proposed by Olivecrona et al. \cite{olivecrona2017molecular} uses RL to tune the MLE pre-trained RNN on the molecular de-novo generation task. It defines the reward as the distance between an augmented likelihood and the agent's likelihood. The augmented likelihood adds a reward of the desirable properties of a molecule onto the log-likelihood of data distribution.
\cite{simm2020reinforcement} computes reward based on fundamental physical properties such as the energy, which is approximated by  fast quantum-chemical method.
Blacshke et al. \cite{blaschke2020memory} augment the REINVENT \cite{olivecrona2017molecular} model with a memory to cope with the mode collapse problem. 
The agent is penalized for generating similar compounds to the ones in the memory unit. 
Atance et al. \cite{atance2022novo} propose the best agent reminder (BAR) loss for training by motivating the agent to update the gradient towards the best agent collected during the training process. It balances between the best agent loss and the REINVENT loss with a factor. The reward function considers the average size of the molecules, drug-like metrics and special molecule DRD2 metrics.
AHC \cite{thomas2022augmented} combines the REINVENT loss with samples selected by the Hill-Climb method \cite{brown2019guacamol}.
Brown et al. \cite{brown2019guacamol} propose an evaluation framework for de-novo molecular design that includes two benchmarks, one for an in-distribution learning test and the other for goal-directed benchmark.
\cite{zholus2024bindgpt} generates 3D molecular conformations by pre-training and fine-tuning a language model, eliminating the need of external
software for graph reconstruction. The state is the current molecular graph and 3D conformations. The action is generating new parts of the molecular graph and 3D conformations. The reward evaluates the binding affinity of target molecular.

Most works above ignore the resolution of molecular design. Recent works start to combine reinforcement learning methods to representations of different resolutions, like fragments \cite{chen2021fragment}, tree \cite{ishitani2022molecular}, and population of candidate molecules\cite{fu2022reinforced}.
FaST \cite{chen2021fragment} investigates PPO \cite{schulman2017proximal} on fragment-based molecular optimization. The action for the agent is to add or delete a fragment from the current molecular.
The reward is $+1$ if the novel qualified molecule is discovered and $-0.1$ if an invalid molecule is explored.
RTJ-RL \cite{ishitani2022molecular} proposes to apply PPO \cite{schulman2017proximal} on a reversible junction tree (RJT) that is a new representation for molecules. RJT representations are convertible to valid molecules, which describe the state of the agent. The action is the modification of the tree, containing information about node, word, site, and stop. 
RGA \cite{fu2022reinforced} combines genetic algorithms and reinforcement learning to optimize the structure-based drug design. 
The state is population at a certain step generation, including the candidate molecules and their 3D poses. Action space is based on crossover and mutation, two main steps in the evolution process. The action is composed of probabilities to choose candidates or ligands in the population.

For molecular design, it is common to use multiple constraints or goals to guide the model. Therefore, multi-objective optimization is also an interesting direction , including weighted sum \cite{goel2021molegular,hu2023novo}, alternation \cite{goel2021molegular,hu2023novo}, reward weighted sum \cite{hu2023novo}, and Pareto optimization \cite{sun2022molsearch,liu2023drugex}.
The state and action of MoleGuLAR \cite{goel2021molegular} are defined as sequence generation. The reward is calculated by solute property measure LogP, drug-likeness metric (QED), and impact in human factor (TPSA). when conflicts exist between rewards, they set all rewards as 0 to guide the generation model towards molecules where the single property is optimal.
Hu et al. \cite{hu2023novo} use the REINFORCE method for fine-tuning. 
They devise a reward-mixing strategy for multiple objective conflicts.
MolSearch \cite{sun2022molsearch} proposes to use MCTS on multi-objective molecular generation and property optimization. The model maintains a global pool for Pareto molecules which are defined as molecules that have at least one property at the best state. 
DrugEx3 \cite{liu2023drugex} uses a Transformer model for the generation to allow users to input prior information like the desired scaffold.

\subsubsection{Reaction Optimization}
It is also feasible to use RL searching in the formula space for chemical applications.
Zhou et al. \cite{zhou2017optimizing} propose to use DRL for chemical reaction optimization. It models a chain of reactions where an agent has the experimental conditions as states and can change the conditions by actions such as increasing the temperature. Once an action is applied, the condition of the reaction changes, and then the agent is required to further take the following action. The reward is about the output of the reaction, such as product yield, selectivity, purity, and cost. RNN as the model is used to construct the agent.
\cite{volk2023alphaflow} uses a similar formulation for shell-growth of core-shell semiconductor nanoparticles.
Gottipati et al. \cite{gottipati2020learning} integrate forward synthesis into reaction selection with modified policy gradient method. The state represents the current molecules generated from a sequence of commercially accessible reactions. The action is the next chemical reaction. Reward is designed to measure the desired properties of generated molecules, such as hydrophilicity and lipophilicity.
\cite{rajak2021autonomous} applies REINFORCE to control threshold temperatures and chemical potentials critical for initiating chemical reactions. The state is current reaction conditions. The action is modifying the the synthesis parameters like temperature and gas concentration. The reward promotes the generation of the target chemical.
\cite{powell2020real} proposes to incorporate multi-agent DDPG into the traditional real-time traditional framework, enabling the optimization to be both economic and adaptable for reaction optimization. The reward aims to optimize yield and efficiency and minimize costs.

\subsubsection{Micro-structure Generation}
\cite{nguyen2022synthesizing} uses a GAN-based SAC algorithm to support generation of 3D microstructure of porous media. The state space contains current design parameters, current QoI (physical quantities of interests) and target QoI. The action is a series adjustments of design parameters. The reward is employed to enhance the alignment of generated microstructure and a target one.

\subsubsection{Quantum Architecture Design}
\cite{kuo2021quantum} propose to do a quantum architecture search by reinforcement learning. The state is defined as a multi-qubit entangled state, the action space is the quantum gate for the design, and the fidelity of the target state measures the reward. The experiment is carried on a simulation environment, which is adapted towards the OpenAI Gym \cite{brockman2016openai}.
\cite{lin2020quantum} designs quantum adiabatic algorithms by DQN networks. The state is the current progress of the adiabatic evolution path. The action is changing the parameters of the numerical operators of energy in the  system. The reward is the success probability of the algorithm.
\cite{ostaszewski2021reinforcement} optimizes Q-learning for quantum circuit architecture. The goal is to control the energy obtained from a circuit and reduce the depth of the circuit depth.
\cite{moro2021quantum} exploits DQN to approximate single-qubit unitary operations with sequences of quantum gates from a finite universal set. The state is the current composition of quantum circuit. The action is selecting a gate from a predefined finite universal set. The reward considers the difference between the target and the generated sequence for fidelity and the length of the sequence.

\subsection{Recommender System and Information Retrieval}
We include most works in this survey about how to generate content without collaboration with humans. There are also applications where RL can be used to generate interactions between two entities, such as a robot and an environment. 
For the recommender systems, it is better for readers to read surveys about how reinforcement learning is injected into the process \cite{chen2021survey,wang2023causal}. In general, user interaction history is considered as states and items are defined as actions. The agent is required to generate items that users might be interested in. Thereby, the user feedback can be incorporated as rewards to guide the learning process. 

\subsection{Robotics}
Robotics is another useful area that deviates from the generative applications mentioned above. Generally speaking, robotics can be treated as an interactive agent that generates responses to humans' orders. Recent advancement \cite{huang2023voxposer} shows that large language models plus visual foundation models might lead to a large step towards better robotics control policies. \cite{wang2023voyager} even proposes a LLM-based agent whose action is generated code to play MineCraft.

\subsection{Other Areas}

Generation models have wide applications. We also list works that apply RL in niche areas like procedure generation\cite{khalifa2020pcgrl}, simulated robotics \cite{ha2019reinforcement}, graph generation\cite{darvariu2021goal}.
PCGRL \cite{khalifa2020pcgrl} proposes an abstract description of formulating the procedure generation problem into an MDP. The procedure generation problem is often used for game construction. They highlight three types of representations for the state and action modelling. Narrow representation only changes game elements at predefined locations. Turtle representation provides an agent with the ability to move around on the map. The broad representation enables an agent to change game elements at other positions.
Ha \cite{ha2019reinforcement} proposes to not only optimize a policy for a bipedal walker to complete tasks but also optimize the shape of the walker. It conducts experiments on the OpenAI Gym \cite{brockman2016openai} BipedalWalkerHardcore-v2 task and learns to generate an agent that can achieve better performance.
Apart from knowledge graphs and causal graphs, reinforcement learning agents can also generate other types of graphs like road graphs and power grid graphs \cite{darvariu2021goal}, where the state is the set of nodes and edges, the action is the selection of a node either the start node or the end node. The reward is defined for robustness study expected critical fraction of nodes to the removal. The reward is estimated by Monte Carlo sampling. Q-learning is employed for this task.
Reinforcement learning is also used in quantum computer design \cite{kuo2021quantum}.
Exploratory Data Analysis (EDA) is another area for RL-based generation. \cite{bar2020automatically} proposes to generate notebooks during EDA by formulating the generation process in MDP. The action space consists of different data operations like filter, group, and backtrack to last operation. The state space is a vector comprises three steps' result data including summarization features like the number of distinct values, the number
of null values, the number of groups, and the groups’ size mean
and variance. The reward is formulated with three terms: interestingness, diversity, and coherence.


\section{Challenges and Future Directions}
Despite benefits and wide application of RL-based methods, challenges also exist, where potential new research directions emerge, opening new opportunities to further improve current generative models and applications. In this section, we introduce a range of challenges, the responses from the community, and less explored promising future directions.

\begin{table} 
\centering
\caption{Challenges and future directions of reinforcement learning methods applied in generative model and applications}
\begin{tabular}{cc}
    \toprule
    Challenges and Future Directions & Related Works \\
    \hline
    Peaked Distribution & \cite{Choshen2020On,kiegeland2021revisiting,ramamurthy2022reinforcement, donato2022mad, honda2023switching} \\
    Exploitation-Exploration & \cite{silver2016mastering, jin2020multi}\\
    Sparse Rewards & \cite{guo2022efficient,anuchitanukul2022surf,korshunova2022generative,yao2020imitation} \\
    Long-term Credit Assignment & \cite{vezhnevets2017feudal,guo2018long}\\
    Generalization & \cite{zhao2023multi} \\
    \bottomrule
\end{tabular}
\label{tab:challenges-future-directions}
\end{table}

\textbf{Peaked Distribution}
Given the fact that reinforcement learning improves the performance of generation for specific objectives, Choshen et al. \cite{Choshen2020On} inquire about the reason behind the performance increase in neural machine translation. 
They propose that the peakiness of the distribution is the true factor for RL to improve performance instead of reward learning. The peakiness illustrates that RL fine-tuning tends to increase the probability of the best answer to make the distribution lower entropy and more discrete. 
The following work \cite{kiegeland2021revisiting} refutes \cite{Choshen2020On} and show that BLEU increase is not tied to the peakiness in RL training. Other subsequent papers methods to address this problem by multi-temperature sampling \cite{dognin2021regen} and finetuning \cite{honda2023switching}.
More work could be conducted to improve the diversity of the generated results in specific areas. Potential solutions may be found in literature of popularity bias \cite{honda2023switching}.

\textbf{Exploitation and Exploration}
In reinforcement learning, an agent must balance a trade-off between exploitation and exploration. To maximize the expected reward, the agent must exploit the best action of what it has experienced before. But to collect the potential best action for learning, it must explore broadly all possible situations to gather enough data in the learning process. This dilemma is not suited for offline settings such as classic supervised learning and unsupervised learning.
One solution for exploration is Upper-Confidence-Bound (UCB) action selection, which injects the consideration of exploration by the times of selection for action. UCB has been applied in molecular design \cite{chen2021fragment}.

For sequential generation problems, action space is prohibitively large which makes the exploration difficult. Therefore, pretraining-finetuning architecture is widely exploited to alleviate this problem.
Another way to tackle this problem is searching in the action space and composing meta-action to decrease the difficulty of exploration. Similar to AlphaGo \cite{silver2016mastering}, RationaleRL \cite{jin2020multi} proposes to employ an MCTS in searching a relatively small and important action space to improve the performance of the RL-based generator.
A new recent approach of exploring the solution space is applying MCTS in the inference time of Large Language models \cite{zhao2024large,zhang2024rest}, which opens new direction that self-play could enhance the performance of current LLMs.

\textbf{Reward function design and multi-objective optimization}
A large number of works have been conducted on how to guide model training with hand-designed new signals by RL. Multiple objectives are usually utilized for various constraints and guidance modelling. It is ideal that the optimal values can be achieved all at once. But it is often not the case, making the Pareto optimization a useful tool for analyzing the result. While according to our exploration, few works have addressed multi-objective optimization and Pareto optimization in application areas. It might be valuable to trade-off in multiple contradictory losses. Therefore, it is advantageous to explore how to trade off between them for a more robust model.

Another problem is \textbf{sparse rewards}.
It is easy to see that rewards are training signals that shape the action pattern of agents. In an ideal environment, rewards are emitted at each action to provide sufficient guidance for models. If one reward is computed based on the generated sequence \cite{guo2022efficient,anuchitanukul2022surf}, the environment must produce a reward at the end of each episode. Some reward evaluation such as BLEU \cite{papineni2002bleu} has this problem. Rewards become sparse and pose challenge for the learning model to tell which action has a higher impact on the last rewards.
Guo et al. \cite{guo2022efficient} proposes a multi-step path consistency learning to address this problem by taking a sequence as a whole, where the consistency is computed over multiple steps instead of single step to tackle the sparse reward problem.
SURF \cite{anuchitanukul2022surf} defines a new reward function that emits a sequence of normalized rewards computed on incomplete sentences and the target sequence. 
Korshunova et al. \cite{korshunova2022generative} address sparse reward problems by fine-tuning, experience replay, and real-time reward shaping. The fine-tuning and experience replay could help the policy focus on promising but rarely explored molecules, and reward shaping provide dynamic process rewards.
Another way to tackle the sparse reward issue is introducing the label from human experts as a new signal \cite{yao2020imitation}, which incorporates a DAgger-like method into semantic parsers that interact with users by texts.
DAgger \cite{ross2011reduction} is an imitation learning method that works by collecting new data from experts and integrate them into the old dataset, which is used for training a new policy.
Despite mentioned methods, Similar ideas like reward-shaping could be applied according to specific task configuration, which is also a promising direction to further enhance the model.

\textbf{Long-term Credit Assignment}
The credit assignment problem arises because the agent need to determines which previous actions have impact on current rewards. With the increase of time steps, it becomes more difficult for the increase  of the number of previous actions.
Hierarchical RL is introduced to cope with this problem\cite{vezhnevets2017feudal}. It sets up two modules for learning, a manager and a worker. 
The manager takes the latent representation and produces a goal vector in a low-dimensional space. The worker fuses the latent vector and the goal to make the decision. 
LeakGAN \cite{guo2018long} uses an agent with hierarchical structure for image captioning to deal with the long text generation problem. The hierarchical architecture is similar to \cite{vezhnevets2017feudal}. The difference is that policy gradient algorithm is introduced into a GAN network, where reward comes from the discriminator in GAN.
With the increase of model capability, more difficult and complex tasks are constructed to test the boundary of the generative models. Hierarchical methods could potentially improve the long-horizon problem solving ability for the generative models.

\textbf{Generalization}
Classic reinforcement learning algorithms are often designed for specific tasks without considering task adaptation \cite{hospedales2021meta}. Classic RL models often perform worse on unseen tasks\cite{open2021open}. Meta reinforcement learning like MAML \cite{finn2017model} is designed to adapt agents for meta-learning tasks. The meta-learning setting involves meta-training 
and meta-testing environments, mimicking the few-shot supervised learning setting where at test time, there are a few examples for the algorithm to adapt to the test tasks. In order to adapt to new tasks, MAML learns an adaptive initialization of neural network parameters. 
This design is model-agnostic and can be applied in various machine learning tasks such as supervised learning and reinforcement learning.
It can also be introduced in multi-scenario text generation for adaptation of different scenarios \cite{zhao2023multi}, where Zhao et al. combine MAML with discriminator-guided text generation.

Recent works also show that it might be difficult to generalize logic inference tasks even for the best GPT-4 \cite{liu2023evaluating}. Thereby, the difficulty of generalization of deep generative models might be further addressed by the RL algorithms. More work should be devoted to how to design a model that can better generalize and achieve better results on out-of-distribution data. 
Retraining \cite{tripp2020sample},  and causal machine learning \cite{scholkopf2021toward} might be interesting ways to complement the capability of RL-based generators for better learning and adaptation.

\textbf{Model Enhancement and Control}
Recent research has tailored RL to help solve the difficulty of sampling of EBM via RL as a distributional approach \cite{parshakova2019distributional} or increasing the efficiency by searching backward propagation of DDPM \cite{fan2023optimizing}. These approaches pave a new way to exert RL to improve the generative model, in contrast to classic applications where RL is a way to introduce new training signals or serve as an architecture builder.
The aforementioned works show the feasibility of RL application, more advancement in the RL community might be transferred to these models to further improve the model performance. A good example is \cite{korbak2022on}, which compares the difference between distributional formulation and policy gradient but still exploits the similarity which lay the foundation for variance reduction to be applied to distributional methods.

\textbf{Human Alignment in LLM and foundation models}
LLM has demonstrated great transferability on a number of sequence modeling problems. Combined with vision foundation models, it is a potential way to achieve more powerful models for diverse task settings. The fast-developing literature in this area is drawing the capability of current large models and exposing new problems for methods like RLHF. This might foster new directions for how RL is incorporated into generative models.


The RLHF method \cite{stiennon2020learning} is a hot research direction since the high impact of large language models \cite{openai2022chatgpt}. New studies also emerge to explore whether the RL is an indispensable factor. Preference-based models have been proposed to model human preference. Direct Preference Optimization (DPO) \cite{rafailov2023direct} aims to substitute reinforcement learning by directly utilizing reward functions by preference modelling. More research can be conducted in this line to find an optimal way for preference modelling.
Also, human preference also is dynamic, so the capture of dynamics and improvement on current generative applications might be an interesting direction.


\section{Conclusion}

In this survey, we propose a unified taxonomy for RL applied in generative AI. We collect works from various directions and extract the key usage of reinforcement learning. We first briefly introduce the concept of generative models and reinforcement learning methods. Then we introduce the key application methods for RL to be incorporated into the generative models. Furthermore, we extract exemplar works in a range of application areas for readers who want to narrow down to a specific area. Finally, we show the promising directions of future research and conclude the whole survey.


 
\bibliographystyle{IEEEtran}
\bibliography{bare_jrnl_new_sample4}
%


 




\vfill

\end{document}